\documentclass[10pt,twocolumn,letterpaper]{article}

\usepackage[pagenumbers]{cvpr} %

\usepackage[dvipsnames]{xcolor}

\usepackage{graphicx}
\usepackage{amsmath}
\usepackage{amssymb}
\usepackage{booktabs}
\usepackage{float}
\usepackage{xcolor}
\usepackage{colortbl}
\usepackage{stmaryrd}
\usepackage{listings}
\usepackage{pgfplots}
\pgfplotsset{compat=1.16}
\usepgfplotslibrary{groupplots}
\usetikzlibrary{patterns}

\usepackage[accsupp]{axessibility}

\usepackage[dvipsnames]{xcolor}
\usepackage[normalem]{ulem}

\newcommand{\PROBLEMNAME}{Illustrated Instructions\xspace}

\newcommand{\METHODNAME}{StackedDiffusion\xspace}

\definecolor{cvprblue}{rgb}{0.21,0.49,0.74}
\usepackage[breaklinks,colorlinks,citecolor=cvprblue]{hyperref}

\usepackage{tabularx}

\usepackage{caption}
\usepackage[capitalize]{cleveref}

\crefname{section}{\S}{\S\S}
\crefname{subsection}{\S}{\S\S}
\crefformat{table}{Table~#2#1#3}
\crefformat{figure}{Figure~#2#1#3}
\crefformat{equation}{Eq~#2#1#3}

\def\confName{CVPR}
\def\confYear{2024}

\title{Generating Illustrated Instructions}

\author{Sachit Menon$^{1,2}$\thanks{Work done during an internship at Meta.} \qquad Ishan Misra$^{1}$ \qquad Rohit Girdhar$^{1}$ \\
$^{1}$GenAI, Meta \qquad $^{2}$Columbia University \\
{\footnotesize \url{https://facebookresearch.github.io/IllustratedInstructions}}
}

\begin{document}

\maketitle

\begin{abstract}
    We introduce a new task of generating ``\PROBLEMNAME'', \ie visual instructions customized to a user's needs.
    We identify desiderata unique to this task, and formalize it
    through a suite of automatic and human evaluation metrics, designed to measure the validity, consistency, and efficacy of the generations.
    We combine the power of large language models (LLMs) together with strong text-to-image generation diffusion models to propose a simple
    approach called \METHODNAME, which generates such illustrated instructions given text as input. The resulting model strongly outperforms baseline approaches and state-of-the-art multimodal LLMs; and in 30\% of cases, users even prefer it to human-generated articles. Most notably, it enables various new and exciting applications far beyond what static articles on the web can provide,
    such as personalized instructions complete with intermediate steps and pictures in response to a user's individual situation.
 \end{abstract}

\begin{figure}[t]
  \includegraphics[width=\linewidth]{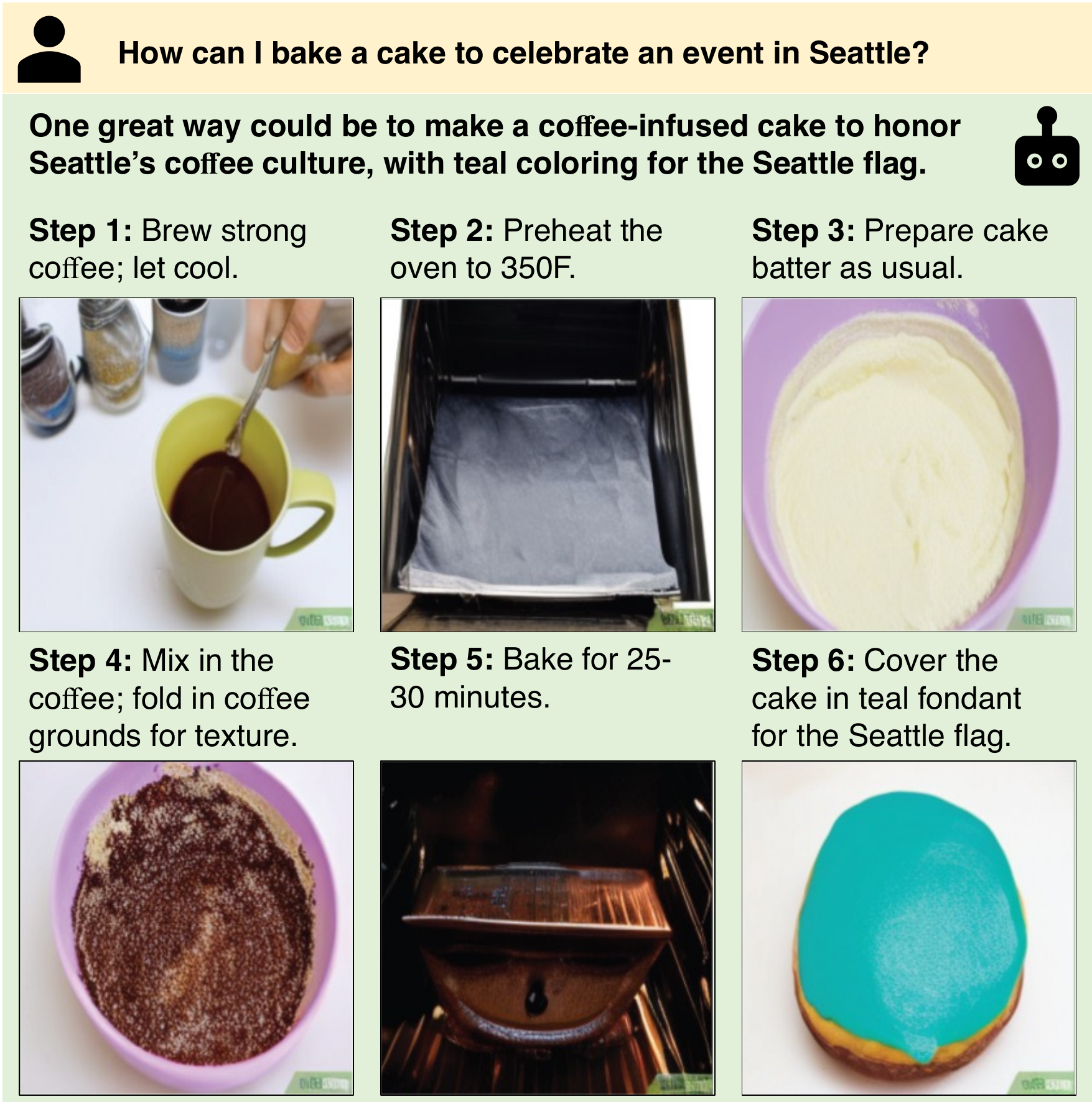}
  \caption{
  \textbf{\METHODNAME generating \PROBLEMNAME.} Given a goal (or any textual user input), \METHODNAME produces a customized instructional article complete with illustrations that not only tells the user how to achieve the goal in words, but also shows the user by providing illustrations.
  }
  \label{fig:teaser}
\end{figure}
\section{Introduction}
\label{sec:intro}

The internet is a vast resource to find answers to all types of questions.
It is often easy to find a webpage or a video that walks through the exact steps to achieve a user's goal.
With the rise of Large Language Models (LLMs) trained on internet-scale data, users can just ask the LLM for instructions to achieve a goal.
This allows the users to get answers for specific personalized queries for which there may not be an existing webpage on the internet, \eg, modifying cooking recipes with user specific dietary restrictions.
Moreover, if the users make a mistake when following the instructions, simple follow-up questions to the LLM can generate alternate instructions, a major advantage over static web search.

In spite of such advantages, LLMs still have one major limitation -- they cannot
generate visuals, which are critical for users to learn from and follow instructions for a wide range of tasks~\cite{carneyPictorial2002}.
Consider, for instance, instructions that require visual inspection, \eg, a recipe that requires the user to stir fry onions until golden brown, or searching for bubbles when a flat tire is submerged under water.
An image accompanying such instructions
can make following them %
significantly easier. %
Can we develop methods with the strengths of LLMs that can also generate such visuals?

In this work, we tackle this challenge, developing models that can not only \textit{tell} a user how to accomplish their task, but also \textit{show} them how.
We define the novel task of {\bf \PROBLEMNAME}: creating a set of steps with visualizations that showcase an approach to solve the user's task. We carefully consider the different dimensions of the problem, laying out three desiderata unique to this setting. To measure these desiderata, we develop automated metrics based on prior work in instructional article assessment~\cite{yang2021visual} and image generation~\cite{ruizDreamBoothFineTuning2023}.

We propose a new model to solve the task of \PROBLEMNAME that combines LLMs with a text-to-image diffusion model.
We train our model on stacks of instructional images from websites such as WikiHow. The resulting {\bf \METHODNAME} model creates full instructional articles, complete with customized steps and a sequence of images uniquely generated to describe those steps. It leverages large-scale pretrained LLMs and finetuned text-to-image diffusion models, and employs techniques to accomplish the task without the need to introduce any new learnable parameters.
This includes spatial tiling for simultaneous multi-image generation, text embedding concatenation to reduce information loss in long instructions, %
and a new ``step-positional encoding'' to better demarcate the different steps in the instructions.
Thus, \METHODNAME can %
generate useful illustrations even when trained even with limited instructional data (\cf~\cref{fig:data_ablation}).

We compare \METHODNAME with various baseline approaches based on off-the-shelf tools, and find that they all fall short,
even when used in conjunction with one another.
Existing T2I models are incapable of generating visuals directly from a user query.
Even when given more detailed instructional text, we show that existing T2I models fail to produce images that are simultaneously faithful to the goal, the step, and consistent among each other.
Given the recent introduction of multimodal LLMs~\cite{koh2023grounding,aghajanyan2022cm3}, we also compare with a recent open-source model, GILL~\cite{koh2023generating}, and show such models also fall short of consistently generating useful visuals along with text.
We posit that our approach of leveraging the spatial priors learned by pretrained diffusion models to generate multiple images together, in conjunction with pretrained LLMs, is significantly more compute and data efficient than an approach purely based on next-token prediction, without these priors.
Our thorough ablations and human evaluations show that \METHODNAME convincingly surpasses state-of-the-art models.
Our final model even outperforms \textit{human-generated} articles in $30\%$ of cases, showing the strong potential of our approach. %

{\noindent \bf Contributions:} 1) We introduce the novel task of \PROBLEMNAME, which requires generating a sequence of images and text that together describe how to achieve a goal (\cref{sec:problem,sec:data,sec:quant}), along with desiderata and metrics for this task; 2) We propose a new approach \METHODNAME for \PROBLEMNAME, with novel modifications to the modeling procedure, enabling generation of visuals suitable for instructional articles for the first time without any additional parameters (\cref{sec:method}); 3) We show that our proposed method achieves strong performance on all metrics, and confirm that human evaluators prefer it over existing methods by wide margins--even surpassing ground truth images in some cases (\cref{sec:baselines,sec:ablations,sec:sota}); 4) Finally, we showcase new abilities that \METHODNAME\ unlocks, including personalization, goal suggestion, and error correction, that go far beyond what is possible with fixed articles (\cref{sec:applications}).

\section{Related Work}
\label{sec:relwork}

{\noindent \bf Instructional data, tasks, and methods.} In the text domain, learning language models on  WikiHow~\cite{koupaeeWikiHowLargeScale2018,zhangReasoningGoalsSteps2020}, has led to advances in tasks such as summarization \cite{zhangPEGASUSPretrainingExtracted2019}, commonsense procedural knowledge \cite{zhouLearningHouseholdTask2019,zhangReasoningGoalsSteps2020}, question answering \cite{dengJointLearningAnswer2020}, and hierarchical reasoning \cite{zhouShowMeMore2022}. In particular, \citet{zhangReasoningGoalsSteps2020} introduce the \textit{goal inference} task, in which a model is presented a goal text as input and asked which of 4 candidate steps is one that actually helps achieve that goal, as well as the analogous \textit{step inference} task. %
In the multimodal setting, \citet{yang2021visual} introduce the Visual Goal-Step Inference (VGSI) dataset and task, which consider articles of interleaved text and images. In this task, a model is again presented goal text as input but is asked which of 4 candidate images is one that actually helps achieve that goal. %
They show representations learned on this data aid in tasks related to instructional videos \cite{miech19howto100m,tangCOINLargescaleDataset2019}.

{\noindent \bf Learning representations from multimodal data.} Multimodal data (including multimodal instructional data, such as from video \cite{miech19howto100m,tangCOINLargescaleDataset2019}) has proven to be a powerful source of signal for tasks such as zero-shot recognition \cite{radford2021learning,jia2021scaling}, text-image and text-video retrieval \cite{gabeur2020multi,gabeur2022masking,chen2021multimodal,liu2021hit,wang2021dig,bain2021frozen,luo2021clip4clip,miech2021thinking,luo2020univl}, temporal segmentation~\cite{yang2021taco,sun2019learning,luo2020univl,zhu2020actbert,piergiovanni2021unsupervised}, activity localization~\cite{chen2021multimodal,xu2021boundary,zhukov2020learning,yang2021taco,luo2020univl,linLearningRecognizeProcedural2022}, anticipation~\cite{girdhar2021anticipative,sener2019zero,wu2022memvit,dessalene2021forecasting,furnari2020rolling}, question-answering~\cite{yang2021just,seo2021look,li2020hero}, summarization~\cite{iv-sum,clip-it}, and even recipe personalization \cite{fatemiLearningSubstituteIngredients2023}. %

Existing work on instructional data has centered around \textit{understanding}, rather than generation. We instead focus on the novel setting of generating full multimodal articles complete with text and illustrations.

{\noindent \bf Generative models.}
Recent work has examined text-conditioned visual generation through autogressive~\cite{changMuseTextToImageGeneration2023,yuScalingAutoregressiveModels2022} or diffusion models~\cite{sahariaPhotorealisticTexttoImageDiffusion2022,rombachHighResolutionImageSynthesis2022,balaji2022ediffi,ramesh2022hierarchical,nichol2022glide,gu2022vector,gafni2022make}. These advances have been leveraged to create text and images together with purely autoregressive~\cite{aghajanyan2022cm3,yu2023scaling} or combined approaches~\cite{koh2023generating}. Concurrent work ~\cite{aiello2024jointly,sun2024emu} enables the capability of generating multiple images together in the autoregressive framework, but requires substantial additional parameters and focus on creating images that adhere solely to the nearby text rather than enabling consistency. (Similar issues arise in the text-to-video setting; see Appendix~\ref{sec:appdx:t2v}.)
\METHODNAME, on the other hand, leverages the priors built into the T2I model to obtain consistency without any additional parameters. %

\begin{figure}[t]
    \centering
    \includegraphics[width=\linewidth]{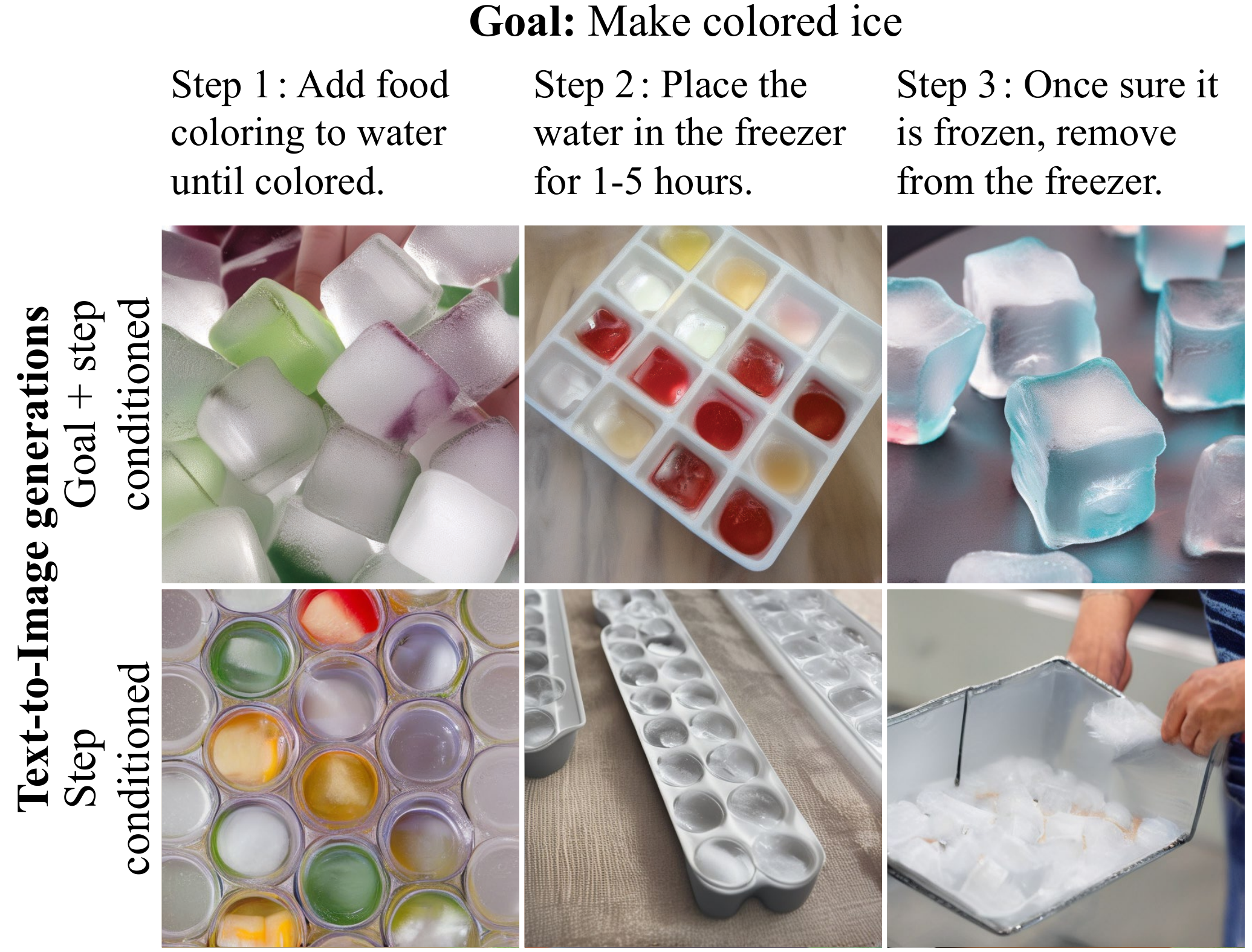}

    \caption{{\bf Failure modes of a naive approach.} A frozen T2I model is not able to capture both the goal and the step, showing only one or the other depending on how it is prompted. Further, it can not produce consistent images, leading to odd changes such as the color of the ice varying between images.}
    \label{fig:basefails}
  \end{figure}

\section{\PROBLEMNAME}
\label{sec:problem}

\label{sec:desiderata}

We now formalize our task and the corresponding desiderata.
The input to the system is a goal text $\bf g$. As output, we would like to produce step text ${\bf s}_i$ as well as step illustration ${\bf I}_i$ for each step $i$.
The step text should be a natural language description of the step, while the step illustrations should be an image corresponding to the step.

On first glance, one might think that this task is a straightforward amalgamation of textual instruction generation and text to image generation.
While these tasks are closely related and necessary substeps towards illustrated instructions, they are not sufficient.
The naive approach--simply creating step text ${\bf s}_i$ from goal text $\bf g$, and each image ${\bf I}_i$ individually from each ${\bf s}_i$--fails to recognize the fresh new challenges that do not exist in either of these two other tasks.
We identify three key requirements for useful illustrated illustrations: goal faithfulness, step faithfulness, and cross-image consistency.

The first requirement is clear: if the images do not relate to the goal, they cannot be good illustrations.
This motivates the desiderata of \textit{goal faithfulness}, which requires that each image faithfully reflects the goal text.

However, goal faithfulness alone is not enough. Consider the first row of images in~\cref{fig:basefails}. The images all reflect the ultimate goal--creating colored ice--but fail in their role as step illustrations. This in turn motivates \textit{step faithfulness}, which requires that each image be faithful to the step text.
We see in this example that baseline T2I models fail dramatically along this metric; every image reflects the goal rather than the specifics of the step requested.

Finally, the generated images should be consistent with each other. While the second row of images in~\cref{fig:basefails} are all faithful to the step text, the color of the ice (and even the style of the images, cartoon or real) changes between images. This is jarring and confusing to the reader.
This motivates the final criterion of \textit{cross-image consistency}, which requires that each image be consistent with the other images produced for a particular generation. %

\begin{figure}[t!]
  \centering
  \includegraphics[width=\linewidth]{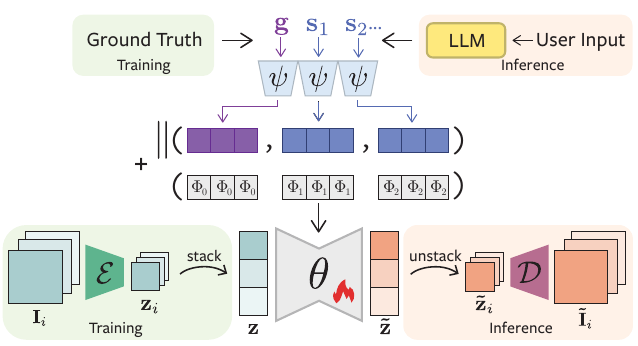}
  \caption{{\bf Overview of \METHODNAME.}
  At training time, we use the given goal and step text, and stack the encoded ground truth step-images. At inference time, we obtain the goal and step text from an LLM, and unstack denoised latents to produce the output images.
  See \cref{sec:method} for details and notation.
  }
\label{fig:method}
\end{figure}

\section{\METHODNAME}
\label{sec:method}

We propose a new architecture, \METHODNAME, to overcome the limitations of existing text-to-image (T2I) approaches for the task of generating interleaved text and images for instructional tasks.
\METHODNAME builds upon T2I models based on latent diffusion models (LDMs)~\cite{rombachHighResolutionImageSynthesis2022}, which are diffusion models~\cite{ramesh2022hierarchical} that operate on a low-resolution `latent' encoding of the original images. %
Our primary desiderata is that the images generated must be faithful to both the goal and step text.
However, in initial experiments we found that simply encoding the goal and step texts joined together (as strings) leads to an uninformative encoding. For instance, generations conditioned on this combined text tended to ignore certain steps or the goal. %
Furthermore, the length of this combined text will likely exceed the context length limitations of the text encoders~\cite{radford2021learning} commonly used in T2I models, leading to undesired truncation and information loss.

Hence, we elect to use a more general approach, applicable to text encoders with any context length.
Rather than encoding the combined goal and step text to obtain the condition, we first separately encode the goal and step texts and then concatenate the encodings.
This allows the model to learn to use the goal and step text independently, as well as in combination. We add a `step-positional encoding' $\Phi_i$ to this concatenation.
It is broadcast across the dimensions pertaining to a particular step, to indicate to the model where a step begins and ends. $\Phi_0$ denotes the positional encoding reserved to indicate the goal embedding. Hence, given a text encoder $\psi$, and $N$ steps in a given goal, we compute the overall text conditioning $\mathcal{C}$ as
\begin{equation}
  \mathcal{C} = \bigparallel \left( \psi \left( {\bf g} \right) + \Phi_0, \bigparallel_{i=1}^{N} \psi \left( {\bf s}_i \right) + \Phi_i \right)
\end{equation}
where $\bigparallel$ is the concatenation operation. This is shown in~\cref{fig:method} (top).
This design decision is critical to obtain good goal and step faithfulness.

In addition, independently generating each image does not achieve the requirement of cross-image consistency as the model cannot exchange information across images.
Thus, we generate all the ${\bf I}_i$ images at once, which allows the model to jointly generate the sequence and achieve cross-image consistency.
The remaining question, of course, is how to accomplish this simultaneous generation %
such that %
it gives rise to cross-image consistency. Our key observation is that T2I models already have a strong prior for consistency within a single image. Could we make use of this previously-learned knowledge?

We propose a simple method to accomplish this: \textit{spatial tiling}.
As illustrated in \cref{fig:method}, the denoising U-Net is given latents ${\bf z}_i$ corresponding to each of the output images simultaneously, tiled spatially as if a single image.
At training time, training images are encoded into the latent space as usual, then reshaped into the tiled format. In detail,
\begin{align}
  {\bf z}_i &= \mathcal{E} \left( {\bf I}_i \right), \qquad {\bf z} = \bigparallel_{i=1}^{N} {\bf z}_i \\
  L_{LDM} &:= \mathbb{E}_{{\bf z},\epsilon\thicksim\mathcal{N}(0, 1),t} \left[ \lVert \epsilon - \epsilon_\theta ( {\bf z}^{(t)}, t, \mathcal{C} ) \rVert_{2}^{2} \right]\label{eq:loss}
\end{align}
where $\mathcal{E}$ is the encoder to  map the image to the latent space (for instance, using a VAE~\cite{kingmaAutoEncodingVariationalBayes2014}), $L_{LDM}$ is the training objective for the LDM, $t$ denotes a timestep of the diffusion process, $\epsilon$ denotes the noise added at a given timestep, and $\theta$ denotes the parameters of the learned denoising U-Net. At inference, we use classifier-free guidance~\cite{nichol2022glide} to generate a latent ${\bf \tilde{z}}$ given conditioning goal and $N$ step texts. We split the latent into ${\bf \tilde{z}}_1, ..., {\bf \tilde{z}}_N$, and decode into generated images using a decoder $\mathcal{D}$ corresponding to the encoder $\mathcal{E}$, \ie ${\bf \tilde{I}}_i = \mathcal{D} \left( {\bf \tilde{z}}_i \right) $.
(We opt for tiling along a single spatial dimension for simplicity, and find that alternative tiling strategies do not significantly affect performance; see Appendix~\cref{sec:appdx:tiling}.)
We refer the reader to the Appendix~\cref{sec:appdx:prelims} and~\cite{rombachHighResolutionImageSynthesis2022} for further details on LDMs.
Given the stacked conditioning and generation operations, we refer to our final model as \METHODNAME.

During training, goal and step text are obtained from ground truth, while at inference, a pretrained LLM is used to transform arbitrary user input text to an inferred goal and generated step texts. We describe the LLM inference procedure and prompt engineering in the Appendix~\cref{sec:appdx:promptengg}. We initialize the U-Net using a pretrained T2I model, and finetune all layers when training with the stacked input.
Our stacked conditioning only increases the spatial resolution of the input to the U-Net for which can be modeled entirely using the existing parameters (spatial convolutions, attention) of the U-Net.
We use the T2I model's text encoder ($\psi$), kept frozen, to encode the goal and step texts.

Since the spatially stacked latent has a spatial resolution comparable to the ones used for high resolution image generation, we encounter issues observed by prior work~\cite{linCommonDiffusionNoise2023} for high resolution training.
Specifically,~\cite{linCommonDiffusionNoise2023} finds that high resolution latents retain substantial information about the input being noised with typical noise addition schedules, even at the final diffusion timestep $t=T$. This results in an unintentional distribution shift between train time (when all observed inputs contain signal) and test time (when the first inputs are pure noise with no signal).
We address this by adjusting the training diffusion noise schedule such that the signal-to-noise ratio (SNR) at the final diffusion timestep $T$ is zero~\cite{linCommonDiffusionNoise2023}.
This ensures enough noise is added during training so as to mitigate this difference between train and test time usage.

\begin{figure}[t]
  \centering
  \begin{tikzpicture}
    \begin{axis}[
        ymin=0,
        area style,
        ylabel={Density},
        axis x line*=bottom,
        axis y line=left,
        label style={font=\footnotesize},
        tick label style={font=\footnotesize},
        height=1.5in,
        width=\linewidth,
    ]
    \addplot+[ybar interval,mark=no,cvprblue,fill=cvprblue!20] plot coordinates { (2, 0.022) (3, 0.21) (4, 0.27) (5, 0.22) (6, 0.14) (7, 0.08) (8, 0.05) (9, 0.03) };
    \end{axis}
\end{tikzpicture}
  \caption{{\bf \PROBLEMNAME data. }
  The histogram shows the distribution of step counts in the data. We find that more than 80\% of articles consist of $6$ or fewer steps.
  }
  \label{fig:data}
\end{figure}

{\noindent \bf Implementation Details.}
We build upon a text-to-image latent diffusion model~\cite{rombachHighResolutionImageSynthesis2022} trained on a large in-house image-text dataset. It leverages a VAE~\cite{kingmaAutoEncodingVariationalBayes2014} to map the images to a $4$D latent space, with a $8\times$ reduction in spatial resolution.
The U-Net largely follows the implementation from~\cite{rombachHighResolutionImageSynthesis2022}, with minor architectural modifications.
As conditioning, it uses the Flan-T5-XXL text encoder~\cite{chung2022scaling}.
We train \METHODNAME for $14000$ steps using the AdamW optimizer \cite{loshchilovDecoupledWeightDecay2019}  with a learning rate of $10^{-4}$, weight decay of $0.01$, and gradient clipping at $\ell_2$ norm of $1.0$. For classifier-free guidance, we use a conditioning dropout rate of $0.05$. We choose to generate at most $N=6$ images simultaneously, because as we will see in~\cref{sec:data}, the vast majority of the data available has $6$ or fewer steps. For shorter training sequences, we pad with empty frames and dummy step texts, and for longer, we drop the extra steps and images. We ablate this choice of $N$ in~\cref{sec:ablations}. Hence in practice, for batch of $8$ sets of $6$ steps at $256$px resolution, we get an encoded latent of $8\times6\times4\times32\times32$. We concatenate the latents spatially into $8\times4\times(6*32)\times32$.
The denoising U-Net is finetuned to denoise these stacked images, leveraging what it has previously learned about spatial consistency, and adapts to this new setting.
At inference time, steps are generated using a pretrained LLM~\cite{openai2023gpt}, prompted to generate at most $N$ steps (\cref{sec:appdx:promptengg}). To generate illustrations, noise is sampled in the shape of the tiled latent with $N$ steps, and the steps generated by the LLM are padded with dummy steps if less than $N$. %
The resulting noise is denoised using the U-Net as usual, and finally reshaped to obtain the $N$ output images upon decoding.
See Appendix~\cref{sec:appdx:implementation} for further details.

\section{Experiments}

\begin{figure}[t]
  \centering
  \includegraphics[width=\linewidth]{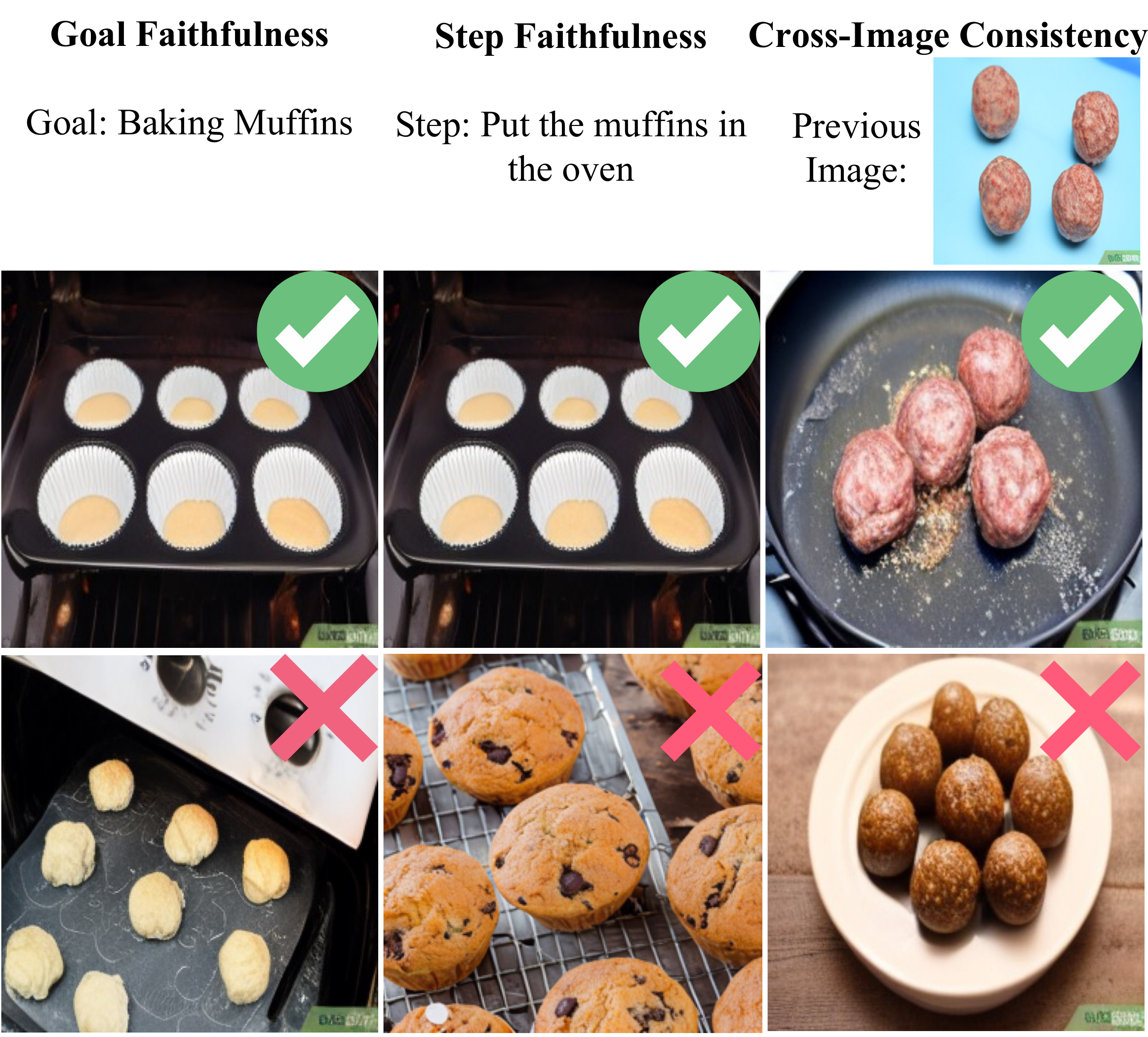}
  \caption{{\bf Metrics.} We introduce three metrics, one for each desideratum presented in~\cref{sec:problem}.
  Goal faithfulness: the second image does not show muffins. Step faithfulness: the second image does not show the step. Cross-image consistency: the second image shows a different number of meatballs with different visuals.
  }
  \label{fig:metrics}
\end{figure}

We train and evaluate \METHODNAME on web-based instructional data. We compare the generations to multiple established baselines and ground truth, using automatic metrics
and human evaluations. We now describe the data, metrics, baselines, and the key results. Finally, we demonstrate some new applications that \METHODNAME enables, showing how it goes beyond standard instructional articles.

\subsection{\PROBLEMNAME Dataset} %
\label{sec:data}
We introduce a new dataset to train and evaluate models for the \PROBLEMNAME task, by repurposing the Visual Goal-Step Inference (VGSI) dataset~\cite{yang2021visual}.
VGSI consists of WikiHow articles, each of which have a high-level goal, 6-10 natural language steps (see~\cref{fig:data}), and associated image illustrations.
We observe that the same data can be used for generating instructional articles. %
It can provide signal for how the goal and step texts should map to output images, and how images should match with each other.

For evaluation, we construct a held-out set from this same data. %
Used directly, however, the data is not well suited as many of the tasks are too high-level to be useful for consistent illustration.
For instance, ``How to Start a Business in North Carolina'' may have steps ``Brainstorm ideas'' and ``Go to the courthouse'' that have no shared visual content.
As such, we filter the data for evaluation to the ``Recipes'' category, which always has a visually clear end state and where illustrations must be consistent with each other. Please see Appendix~\cref{sec:appdx:data} for more details.

\begin{table}[t]
  \centering
  \small %
  \setlength{\tabcolsep}{2.2pt}
  \begin{tabular}{lccccc}
    & Human $(\uparrow)$ & GF $(\uparrow)$ & SF $(\uparrow)$ & CIC $(\downarrow)$ & FID $(\downarrow)$ \\
    \midrule[1.5pt]
    T2I (Frozen) & 22.0 & 92.9 & 43.2 & 51.3 & 69.3 \\
    T2I (Finetuned) & 33.3 & {\bf 78.8} & 52.4 & 51.5 & 53.5 \\
    \arrayrulecolor{lightgray}
    \midrule[0.2pt]
    \arrayrulecolor{black}
    \textbf{\METHODNAME} & {\bf (ref)} & 74.3 & {\bf 61.5} & {\bf 50.7} & {\bf 39.5} \\
    \textcolor{gray}{Ground Truth} & \textcolor{gray}{82.5} & \textcolor{gray}{81.7} & \textcolor{gray}{73.7} & \textcolor{gray}{50.6} & \textcolor{gray}{(N/A)} \\
  \end{tabular}
  \caption{{\bf Comparison to baselines}. Human evaluation is reported as win rate vs our full \METHODNAME model. GF corresponds to goal faithfulness accuracy, SF to step faithfulness accuracy, CIC to cross-image consistency, and FID to Fréchet Inception Distance.}
  \label{table:comparison}
\end{table}

\subsection{Metrics}
\label{sec:quant}

Having established three desiderata in~\cref{sec:desiderata}, we now turn to the question of how to evaluate them. In addition to evaluating the quality of images using FID~\cite{heusel2017GANs}, we propose three metrics, one for each desideratum, that can be used to evaluate the faithfulness and consistency of the generated article. Finally, given the limitations of automatic metrics for evaluating generative modeling tasks~\cite{steinExposingFlawsGenerative2023}, we use human evaluations as our primary metric for overall quality. We illustrate the metrics in~\cref{fig:metrics} and briefly describe them next. Please see the Appendix~\cref{sec:appdx:metrics} for more details.

{\noindent \bf Goal Faithfulness (GF)} measures how well the generated image is associated to the goal text. We evaluate this by
constructing multiple-choice questions (MCQ) as in VGSI~\cite{yang2021visual}. For each generated image, we compare its CLIP similarity~\cite{radford2021learning} with the correct goal text vs the similarity with the texts of three other randomly selected goals. We compute the accuracy of the model in choosing the correct goal text.

{\noindent \bf Step Faithfulness (SF)} measures how faithfully the generated image
illustrates the step it aims to depict.
An image should match the text for the step it was made for, more than other steps. %
We measure this using CLIP similarity and a MCQ task similar to goal faithfulness, where the image should have higher CLIP similarity with the corresponding step text than the other step texts within the same goal.

{\noindent \bf Cross-Image Consistency (CIC)} evaluates
how consistent the generated images for a goal are with each other and penalizes jarring inconsistencies across the images, such as objects changing color or number.
For instance, if a particular set of ingredients are shown for ``gather the ingredients,'' then the same ingredients should be shown for ``mix the dry ingredients.''
We measure this by computing the average $\ell_2$ distance between DINO~\cite{caronEmergingPropertiesSelfSupervised2021} embeddings of the images for each step, as this considers the similarity of visual (rather than purely semantic) features.
Like prior work~\cite{ruizDreamBoothFineTuning2023}, we also found that CLIP features do not work well for image similarity as they are invariant to critical aspects such as color, number of objects, style \etc.
DINO's self-supervised pretraining objective leads to %
features that are sensitive to the visual aspects particular to an image, rather than being invariant to aspects not captured by category or language.

{\noindent \bf Human Evaluation} is finally used to confirm
that the automated evaluation corresponds to actual human preferences. %
We provide fully rendered articles from each model to human evaluators on Amazon Mechanical Turk (AMT), and ask them to choose which article they prefer, or if they are tied. We then consider the win rate as the proportion for which a majority of evaluators selected the given method compared to the articles produced by \METHODNAME. A win rate of $50$ denotes the methods are tied, whereas $<\!50$ denotes the method is worse than \METHODNAME.

\begin{figure}[t]
  \centering
  \includegraphics[width=\linewidth]{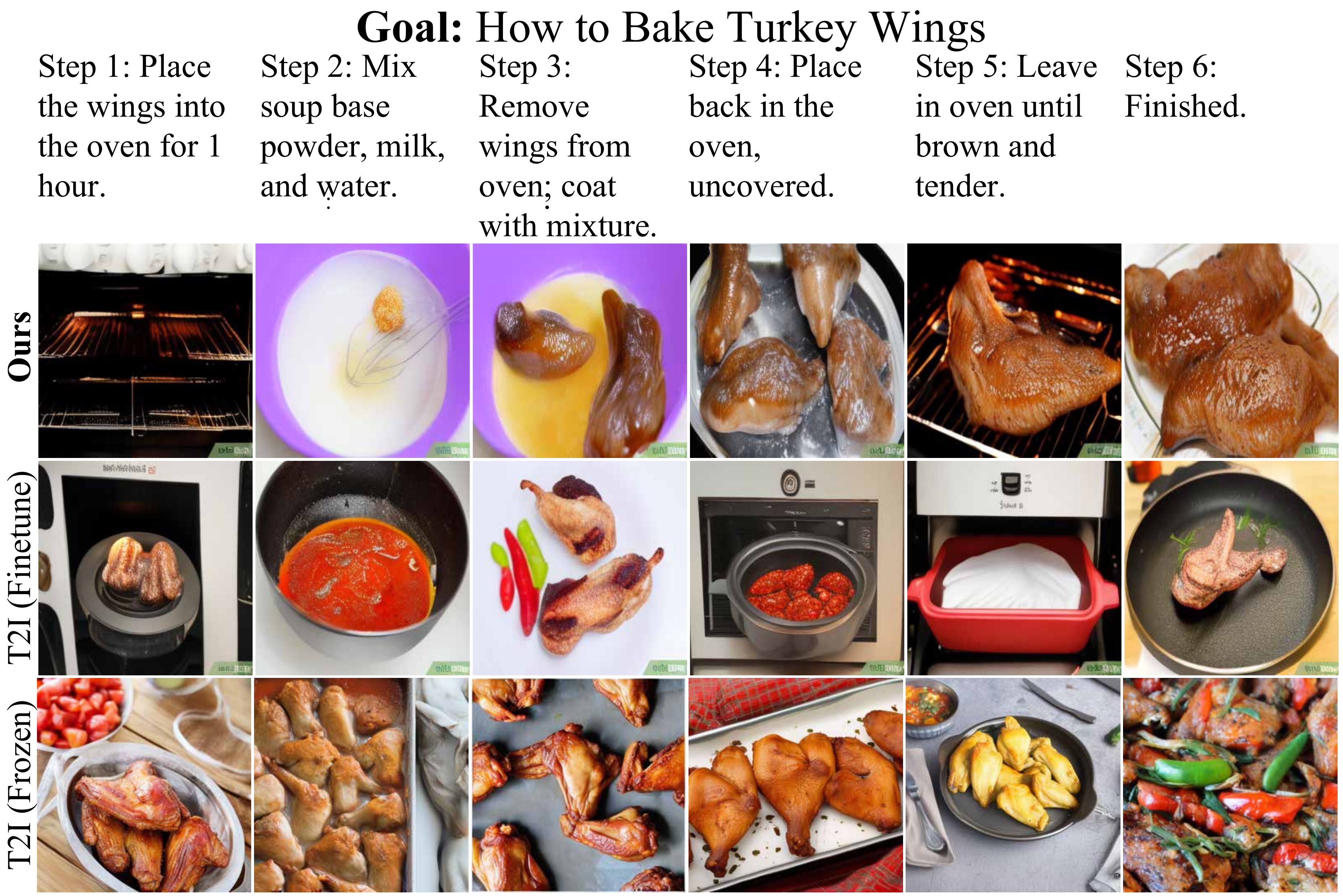}
  \caption{{\bf Baseline comparison.} \METHODNAME images are preferred overwhelmingly over baselines. The frozen baseline tends to only produce images showing the goal, while the finetuned baseline produces images that are more faithful to the step text, but have no visual features in common.}
  \label{fig:vsbase}
\end{figure}

\subsection{Baselines}\label{sec:baselines}
We now describe some baseline approaches based on existing state-of-the-art image generation models, and compare them to \METHODNAME in~\cref{table:comparison}.
{\par \noindent \bf T2I (frozen):}
The most obvious choice for a baseline is to simply use a pretrained and frozen text-to-image (T2I) model. We use our in-house T2I model that is also used to initialize \METHODNAME.
To ensure this baseline gets same information as \METHODNAME,
we prompt the model with a concatenation of the goal with each step, ${\bf g} \bigparallel {\bf s}_i$, for each $i$, and produce a single image ${\bf I}_i$. The sequence then is composed of $N$ independent generations.

We find that these images are faithful to the goal text, but fail to be faithful to the step text and lack consistency. The goal faithfulness is substantially higher than even the ground truth (92.9\% vs 81.7\%). This is because the model is not capable of reasoning about what a particular step towards a goal should look like. Instead, it simply generates an image that matches words in the input text, as illustrated in the first row of~\cref{fig:basefails}. This results in all the images for \eg ``Make Colored Ice" showing the actual finished ice rather than any of the
intermediate steps.
In other words, the step text is eclipsed by the goal text.

{\noindent \bf T2I (finetuned):}
We finetune a T2I model with
the goal and step text embeddings concatenated together as the condition, again producing $N$ independent generations.
We find that this model is more faithful to the step text than the frozen T2I model, but still substantially less than our final model (52.4\% vs 61.5\%). This suggests the goal still overshadows the step text, but to a lesser extent. The goal faithfulness is also lower than for the frozen T2I model, as this model is trained to create images for steps rather than goals. The cross-image consistency is low, similar to the frozen model, as the images are still generated independently.

\begin{figure}[t]
  \centering
  \begin{tikzpicture}
  \begin{axis}
  [
    ybar,
    legend columns = 5,
    legend style={at={(0.5,1.2)},anchor=center,fill=none},
    ylabel={Score},
    axis x line*=bottom,
    axis y line=left,
    ymin=45,
    symbolic x coords={goal, step, cic},
    xticklabels={GF, SF, CIC},
    enlarge x limits=0.25,
    label style={font=\footnotesize},
    tick label style={font=\footnotesize},
    height=1.5in,
    width=\linewidth,
    xtick=data,
    legend image code/.code={
        \draw [#1] (0cm,-0.1cm) rectangle (0.2cm,0.25cm);
    },
  ]
  \addplot [cvprblue!20,fill=cvprblue!20] coordinates {(goal, 72.3) (step, 55.6) (cic, 52.2)};
  \addlegendentry{20\%}
  \addplot [cvprblue!40,fill=cvprblue!40] coordinates {(goal, 73.5) (step, 57.9) (cic, 51.6)};
  \addlegendentry{40\%}
  \addplot [cvprblue!60,fill=cvprblue!60] coordinates {(goal, 74.2) (step, 58.3) (cic, 51.2)};
  \addlegendentry{60\%}
  \addplot [cvprblue!80,fill=cvprblue!80] coordinates {(goal, 74.6) (step, 60.6) (cic, 50.9)};
  \addlegendentry{80\%}
  \addplot [cvprblue,fill=cvprblue] coordinates {(goal, 74.3) (step, 61.5) (cic, 50.7)};
  \addlegendentry{100\% data}
  \end{axis}
\end{tikzpicture}
  \caption{{\bf Effect of training data.} We find that having more training data improves \METHODNAME{}'s faithfulness and consistency. However even with 20\% data it performs well, thanks to its effective use of pretrained T2I model weights.
  }\label{fig:data_ablation}
\end{figure}

\METHODNAME %
achieves substantially improved step faithfulness and cross-image consistency, while maintaining high goal faithfulness. We show an illustrative example in~\cref{fig:vsbase}. Unlike the baselines, the cross-image consistency is %
very close to that of the ground truth data; as generating them together allows for influence between them during the generation process. We also find that the FID of our generated images is substantially lower than that of other models, indicating that our generated images are more similar to the ground truth data. Finally %
on human evaluations, \METHODNAME clearly outperforms them both by a preference of 78\% and 66.6\% respectively. Perhaps most notably, when compared to the ground truth images, human evaluators still picked \METHODNAME 18.5\% of the times, suggesting that our model generates some illustrations even better than the ground truth data. %
Further baselines can be found in Appendix \cref{sec:appdx:baselines}, with a comparison of training and inference costs in Appendix~\cref{sec:appdx:costs}.

\subsection{Ablations}\label{sec:ablations}

{\noindent \bf Step Count.}
As discussed in~\cref{sec:method}, we elect to use a maximum step count of $6$ as it allows sufficient data coverage and most articles fall under this value. We find that human evaluators prefer this model to the shorter step count model of length 4. It is slightly preferred over the model trained with a longer step count of 8, likely as there is insufficient data of long lengths, limiting any advantage that might be gained in longer length generations.
\begin{center}
  \begin{tabular}{c|cc}
    Baseline & 4 Step $(\uparrow)$ & 8 Step $(\uparrow)$ \\
    \midrule[1.5pt]
    6 Step & 32.1 & 46.9\\
  \end{tabular}
\end{center}

\begin{table}[t]
  \centering
  \label{table:gencomp}
  \begin{tabular}{@{}ll@{}@{\hskip 0.1in}|@{\hskip 0.1in}@{}ll@{}}
               & Win Rate & & Win Rate\\
               \midrule[1.5pt]

  T2I (Frozen) & {\bf 33.8} & T2I (Finetuned) & {\bf 34.0}  \\ %
  GILL~\cite{koh2023generating}++         & {\bf 3.9} & LLM + CLIP & {\bf 15.6}     \\
  Goal Retrieval & {\bf 33.1} &   \textcolor{gray}{Ground Truth} & \textcolor{gray}{70.0}   \\

  \end{tabular}
  \caption{{\bf System-level comparison to prior work.} Human evaluation results using fully generated outputs. Reported as human evaluation win rate of the associated  method over \METHODNAME (hence, higher is better for the reported method). \vspace{-1.5em}
  \label{tab:sota}
  }
\end{table}

{\noindent \bf Training data.} In~\cref{fig:data_ablation}, we
randomly sub-sample varying proportions of the total data, and train \METHODNAME on each of those subsets. As the amount of training data increases, it
generates more faithful and consistent images. However even with less data it performs well, owing to its effective use of the pretrained T2I model's initialization.

{\noindent \bf Importance of 0SNR.}
We experiment with the 0SNR technique~\cite{linCommonDiffusionNoise2023} due to the high spatial dimensionality of the latents. We find that not using 0SNR results in substantially reduced step faithfulness, dropping from 61.5\% to 52.8\%.

{\noindent \bf Importance of Step-Positional Encoding.}
We also evaluate the importance of the step-positional encoding,  $\Phi_i$.
This gives the model a critical cue for which parts of the condition correspond to which steps. We find that not using the step-positional encoding results in substantially reduced step faithfulness as well, dropping from 61.5\% to 49.8\%.

\subsection{Comparison to prior work}\label{sec:sota}
The previous metrics were computed with respect to fixed ground truth goal and step texts to provide direct comparisons. However, they do not encompass the full scope of our model's capabilities: the flexibility of the generated text is one of \METHODNAME's core strengths. To evaluate in a closer setting to this real-world usage, we examine the quality of the articles generated by the full system, where the steps are generated with a LLM~\cite{openai2023gpt}. We perform system-level comparisons using human evaluators, including the quality of the generated text and the quality of the images created from the generated text.

We show our results in~\cref{tab:sota}. We first compare to the baselines introduced in~\cref{sec:baselines}, and see results similar to~\cref{table:comparison}. We see that even with LLM-generated text, \METHODNAME outperforms baseline T2I based approaches, either frozen or finetuned, by more than 66\% win-rate.

Additionally, we compare \METHODNAME to
recent state-of-the-art approaches in multimodal generation, as in principle those can also produce instructional text combined with illustrations.
Specifically, we compare to GILL~\cite{kohGroundingLanguageModels2023}--a %
model trained to generate sequences of interleaved text and images. While other similar approaches have been proposed~\cite{aghajanyan2022cm3,koh2023grounding}, GILL is the best open-source model we could access.
Using GILL directly, however, results in a human evaluation win rate of 0\%. We posit this was for three primary reasons. First, the GILL model was unable to generate long enough generations to suffice as an article in comparison to our generations, despite our best efforts in modifying the generation parameters. Secondly, the model often did not produce any illustrations with the text, although we tried various prompting tactics. %
Thirdly, the text or images produced were often not of high enough quality to actually reflect the goal or steps.
We thus introduce an alternative, which we call GILL++. This uses the (superior) text produced by the same language model as in our method, but passes each step text as input to GILL to generate the corresponding image. This results in a human evaluation win rate of 3.9\% against \METHODNAME. See Appendix~\cref{sec:appdx:gill} for details on prompting GILL, and our GILL++ baseline.

\begin{figure}[t]
  \centering
  \includegraphics[width=\linewidth]{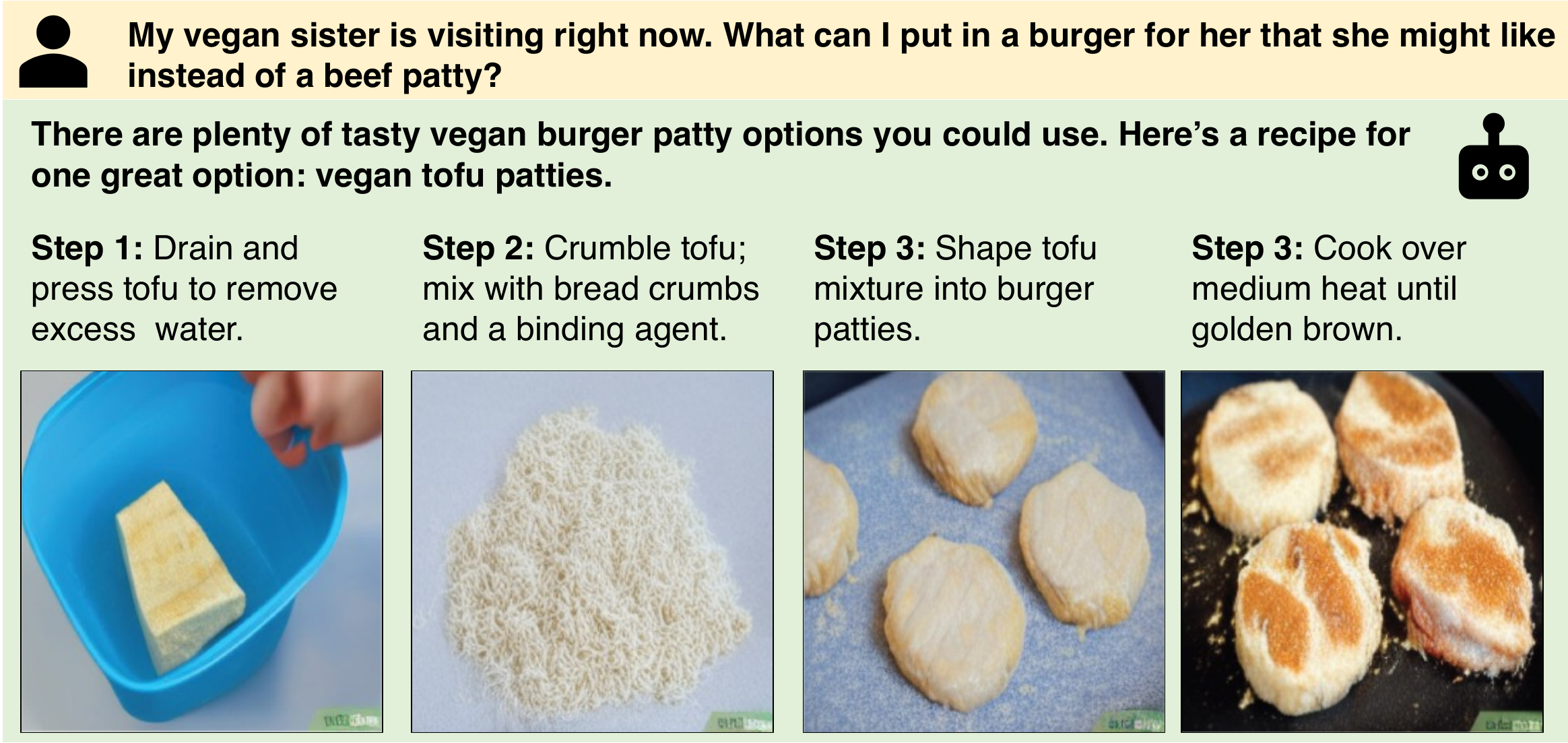}
  \caption{{\bf Personalization.} \METHODNAME enables instructions personalized far beyond what is possible with preexisting articles.}
  \label{fig:personalized}
\end{figure}

Next, we compare \METHODNAME to a retrieval based approach, denoted by LLM+CLIP. Here we retrieve the closest image in the training set according to CLIP similarity to each of the steps created by the LLM (with each concatenated to the goal text, similar to how the frozen T2I model is used). This retrieval-based metric is not suitable for the automated metrics previously introduced as it is based on the same underlying similarity scores as the automated metrics, but we introduce it here for human evaluation. We find that \METHODNAME is preferred overwhelmingly over this retrieval-based approach, despite the retrieval-based approach using real images from the training set. Similarly, we compare to another retrieval-based approach which retrieves the full article with the most similar goal text to the input goal, denoted Goal Retrieval. We find \METHODNAME strongly outperforms this as well. These results together suggest that \METHODNAME is able to generate novel instructional articles that one can not simply retrieve from the training corpus.

Finally, we compare \METHODNAME to the ground truth. Surprisingly again, we find that the human annotators pick our generations $30$\% of times even compared to ground truth.
Note that these articles are handwritten and manually illustrated for the purpose of illustrating these goals.
This shows the strong promise of our proposed approach. We believe with the rapid improvements in training data and generative modeling techniques, future iterations of our model could be indistinguishable or even better than manually created illustrated instructional articles.

\begin{figure}[t]
  \centering
  \includegraphics[width=\linewidth]{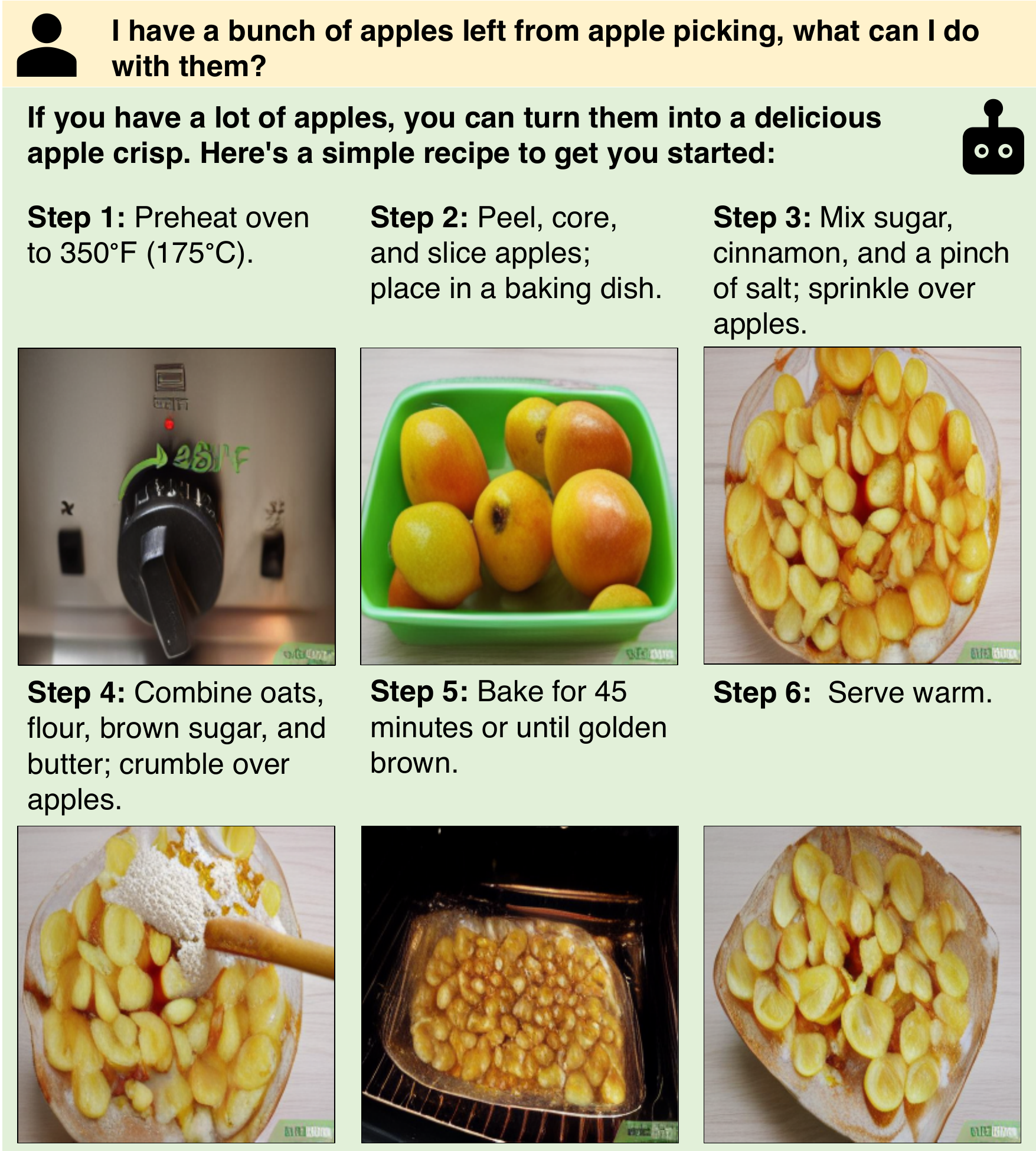}
  \caption{{\bf Goal suggestion and inference.} \METHODNAME can suggest what goal might be most relevant given other information.}
  \label{fig:suggestion}
\end{figure}

\subsection{Applications}\label{sec:applications}
Perhaps the biggest strength of a generative approach is its ability to handle unique user queries that may not fit into the boundaries posed by static articles on the web.
Hence, we demonstrate the capabilities of the full system, combining the strengths of text generation capabilities from the language model with \METHODNAME.

{\noindent \bf Personalized instruction.} A user can provide any situational information specific to their circumstances and obtain an article customized to that situation. This is not possible with fixed, existing articles. For example, in \cref{fig:personalized}, the user can specify a diet and obtain an article for the food they want that adheres to that diet.

{\noindent \bf Goal suggestion.}
As \METHODNAME accepts flexible input text, a user can provide higher-level information about what they want to do and obtain an article with a suggested specific goal that matches what the user would like to do. In \cref{fig:suggestion}, the user describes their situation (having many apples) and \METHODNAME suggests a goal that matches their situation (making apple crisps).

\begin{figure}[t]
  \centering
  \includegraphics[width=\linewidth]{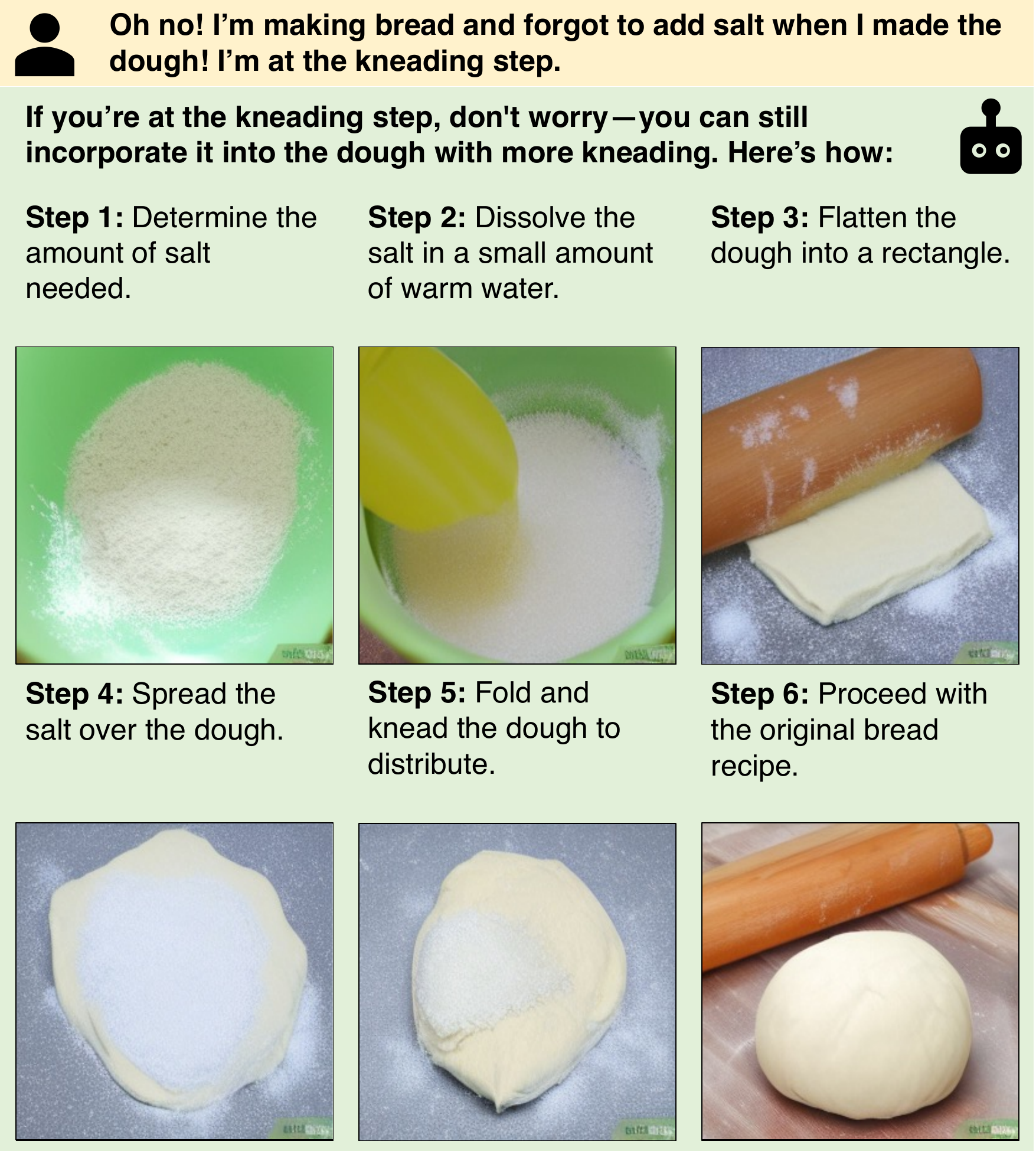}
  \caption{{\bf Error correction.} \METHODNAME provides updated instructions %
  in response to unexpected situations, like a user error.}
  \label{fig:correction}
\end{figure}

{\noindent \bf Error correction.}
Oftentimes, a user in the course of performing a task may make a mistake. Fixed articles provide no avenue for recourse. However, \METHODNAME can adapt to user error and create alternative instructions that best correct for and accomodate the user's situation. See \cref{fig:correction} for an example of this. A fixed article would not provide any alternative paths that adapt as the user performs the action.
This highlights the new avenues opened up by our generative approach to this task.

\section{Limitations and Conclusions}

While \METHODNAME achieves strong performance in generating \PROBLEMNAME, we note that there still remains a gap with ground truth data. Future work might examine how to obtain instructional data at greater scale and how this influences \METHODNAME. Leveraging the latest T2I architectures to improve the faithfulness and quality of our generations would be another promising avenue for future work.
Finally, leveraging improvements in text-to-video techniques to generate video clips describing each step would further improve the usability of such a system.

{\small
\bibliographystyle{ieeenat_fullname}
\bibliography{refs_clean}
}

\clearpage
\setcounter{page}{1}
\maketitlesupplementary

\section{Related Work: Text-to-Video Generation}\label{sec:appdx:t2v}

Works in the text-to-video setting~\cite{singerMakeAVideoTexttoVideoGeneration2022,hoImagenVideoHigh2022,blattmannAlignYourLatents2023,wu2023tune,esser2023structure,hong2022cogvideo,zhou2022magicvideo,luo2023videofusion,Yang2023ProbabilisticAO} are similar to ours in generating multiple images (frames) together, but are typically limited to short time scales, where visual content only changes in minor ways from frame to frame. Some recent works aim to generate longer videos \cite{yinNUWAXLDiffusionDiffusion2023,villegasPhenakiVariableLength2022}, however at a cost of %
substantial parameter overhead over their base T2I models.  %
Many of these methods also use architectural design choices that are specific to this setting, where there is not substantial change from frame to frame, such as factorized spatial and temporal attention; this does not suit our setting, where there are much larger changes across consecutive images.

\section{Tiling Comparison}\label{sec:appdx:tiling}
While the attention operations in our architecture are capable of incorporating context across the full sequence irrespective of the stacking orientation, one might hypothesize that convolutional operations may not have a wide enough receptive field to obtain the same results with different tiling strategies. We experiment with a closer to square orientation for comparison to evaluate this hypothesis. Specifically, we compare to 8x6x4x32x32 $\rightarrow$ 8x4x(3x32)x(2x32)
stacking (closer to square), %
and find the results are indeed similar. In particular, human evaluation gives a win rate of $45.9\%$ for the (3x2) orientation, near even; GF is slightly higher (79.3), SF is slightly lower (60.7), and CIC is marginally higher (50.5). This suggests that the tiling strategy does not have a substantial impact on the model's performance.

\section{Latent Diffusion Models Preliminaries}\label{sec:appdx:prelims}

Latent diffusion models \cite{rombachHighResolutionImageSynthesis2022} model a data distribution in the latent space of another model, such as a VAE \cite{kingmaAutoEncodingVariationalBayes2014}. The latent diffusion model is trained to generate samples from the data distribution $p(x)$ in the latent space, induced by the encoder $\mathcal{E}$, via gradual denoising. These samples can then be decoded by the VAE decoder $\mathcal{D}$ to produce samples from the approximated data distribution. The objective matches the form of our~\cref{eq:loss}.

\section{LLM inference prompt engineering}\label{sec:appdx:promptengg}

We find that simple prompting strategies suffice to have the LLM produce outputs in the format we want. Specifically, we use the following prompt:

\begin{lstlisting}[breakindent=0pt]
{INPUT_TEXT} Write your response in the form of a goal "Goal: {GOAL}" followed by concise numbered headline steps, each one line, without any other text. Use at most 6 steps.
\end{lstlisting}

where \{INPUT\_TEXT\} is any user input, such as those shown in the figures. If only a goal is provided, we provide the input text as ``How can I \{GOAL\}?'', for example ``How can I make colored ice?''.

\section{Data}\label{sec:appdx:data}

We repurpose the VGSI dataset \cite{yang2021visual} for generative purposes. The original dataset is split by step-image pairs. We note that this splits images from a single goal across training and validation. We thus recombine the data, group by associated goal, and then create new splits. Each data point we create is formed by a string of goal text as well as varying numbers of pairs of step text strings and associated images. The distribution of the number of steps is seen in Figure \cref{fig:data}. We find that the overwhelming majority of articles contain less than 6 steps. The training set consists of 95328 goals for a total of 476053 step-image pairs. The validation set, after filtering only for goals in the `Recipe' category, consists of 1711 goals for 9473 step-image pairs.

\section{Metrics}\label{sec:appdx:metrics}
{\noindent \bf Goal Faithfulness (GF). }
We draw on the literature from Visual Goal-Step Inference to define a metric for goal faithfulness. The intuition is as follows. An image should, typically, be more associated with the text for the goal it was made for than for the text of other goals. We can measure this association by computing the cosine similarity between the image and the goal text with pretrained contrastive vision-language models such as CLIP \cite{radford2021learning}. However, CLIP scores are notably miscalibrated when comparing across different image-text pairs. Thus, to make the metric meaningful, we must compare the similarity scores to the similarity with other goals. We accomplish this by constructing multiple-choice questions as in VGSI \cite{yang2021visual}. For each image, we compare the CLIP similarity score with the correct goal text to the CLIP similarity scores with the texts of three other randomly selected goals. We then compute the accuracy of the model in choosing the correct goal text. %
While VGSI uses this metric with a fixed dataset to evaluate their vision-language models, we instead fix the vision-language model used in order to evaluate the data being generated.

{\noindent \bf Step Faithfulness (SF).}
As mentioned in \cref{sec:desiderata}, no matter how well an image reflects the overall goal, it is useless if it does not illustrate the step it serves to illustrate. Inspired by Goal Faithfulness, we can define a similar metric for step faithfulness. An image should be more similar to the text for the step it was made for than for other steps. In particular, it should be more similar to the text than other steps with the same goal. %
It is worth noting that the step faithfulness metric varies in magnitude depending on the number of steps ($N$) in the sequence. As the number of steps increases, the chance that any individual step will have a visual that is not strongly associated with the caption increases, resulting in lower values. However, for a fixed $N$, comparison between models is meaningful.

{\noindent \bf Cross-Image Consistency (CIC).}
Finally, a key component of the \PROBLEMNAME task is that a set of images is generated rather than a single image. These images are not simply independent. If a particular set of ingredients is shown for ``gather the ingredients,'' then the same ingredients should be shown for ``mix the dry ingredients.'' More generally, we would like to avoid jarring inconsistencies between images in the same sequence, such as objects changing color or number, etc. We can measure this by computing the DINO similarity between the images for each step. %
We use DINO \cite{caronEmergingPropertiesSelfSupervised2021} as the measure of visual similarity over CLIP as it reflects \textit{visual} over \textit{semantic} features, which are more relevant for consistency across images.
This has been noted in work for personalized image generation, such as DreamBooth \cite{ruizDreamBoothFineTuning2023}.
CLIP features, in particular, tend to be invariant to the aspects of the image that are most relevant for consistency, such as color, number of objects, style, and more.

{\noindent \textbf{FID.}} We compute FID with respect to the ground truth dataset -- specifically, the 9473 images of the validation set are used as the reference distribution. We use the clean-fid \cite{parmar2021cleanfid} to avoid common pitfalls in FID computation. 

\begin{figure*}[p]
    \centering
    \includegraphics[height=0.8\pdfpageheight]{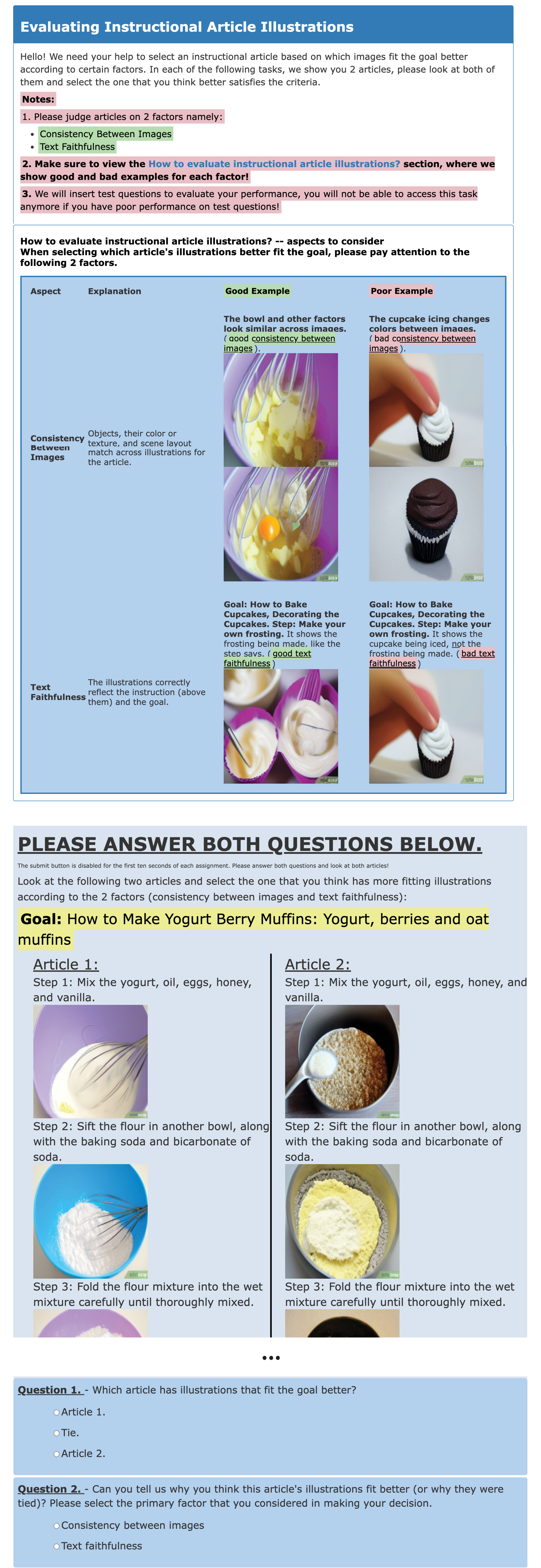}
    \caption{{\bf Human evaluation setup.}}
    \label{fig:human_eval}
  \end{figure*}

{\noindent \textbf{Human Evaluation.}} We show the outputs of each model being compared in a two-way comparison side by side, as shown in Figure \ref{fig:human_eval}. For human evaluation, we select a random subset of $140$ goals that are fixed across evaluations from the validation set. We ask annotators to select which of the two articles they prefer (or if they are tied). As in previous work~\cite{girdhar2023emu}, we find that providing criteria for evaluators to consider during evaluation and requiring a justification for the ultimate decision leads to substantially higher quality evaluations. We use 3 annotators per comparison. To ensure quality of annotation, we selected annotators with the Amazon Mechanical Turk ``Master" qualification that had a $>95\%$ approval rate with at least $1000$ prior tasks completed. For each method, we consider how many goals had a strict majority of annotators pick that method's generations. Tied cases are removed prior to win rate computation.  We report the percentage of majority wins for each method. We find that on average $80\%$ of annotators agreed with the decision for a given sample.

\section{Additional Diffusion Baselines}\label{sec:appdx:baselines}
We include comparisons with two additional baselines here for completeness: a model using a joint encoding for the goal and step text, and a model that only uses the step text as conditioning. 

The goal+step joint-encoding ablation obtains high GF ($91.6$), similar to the frozen model that also operates on a joint-encoding, reflecting that the goal text dominates the embedding. Finetuning, however, improves SF ($50.2$) over the frozen model. The CIC increases slightly ($51.6$) as the influence of the goal information on all images is reduced.

The step-only baseline, as expected, yields much lower GF ($59.5$) as it does not have the goal information; but much higher SF ($77.2$). The cross-image consistency (CIC) worsens substantially ($53.1$) as there is no shared goal information to tie the steps together. This tradeoff leads to overall similar performance to the separate-encoding model in human evals, strongly under-performing our final model.

\section{Training/Inference Cost}\label{sec:appdx:costs}

StackedDiffusion needs only 1 node with 8 GPUs to train,
making it easily accessible and reproducible for the broader research community.
Spatial tiling allows every step to be generated in parallel, maximally utilizing the GPU computation power. Hence, in spite of attending to features for all steps at once, StackedDiffusion takes about the same inference time as a single-step model that requires $N$ forward passes to generate all the steps.
To train to convergence, StackedDiffusion
took 9.7 hours, only
marginally higher than
single-step finetuning (7 hours) on 8 A100 GPUs.

\section{\METHODNAME implementation Details}\label{sec:appdx:implementation}

The U-Net used in the in-house model we build on largely follows the architecture used in~\cite{rombachHighResolutionImageSynthesis2022}, with a T5-XXL text encoder.
The output channels of each block are changed to $(320, 640, 1280, 1280)$.
The GroupNorm epsilon value is $10^{-5}$.
The input text condition uses additional projection and attention layers before being fed into the U-Net as in~\cite{rombachHighResolutionImageSynthesis2022}.
Specifically, this consists of an attention block with $4096$ dimensions followed by layer normalization, a linear projection from $4096$ to $1280$ dimensions, and another layer normalization, and a final linear layer from $1280$ to $1280$ dimensions.

\section{GILL Comparison Details}\label{sec:appdx:gill}

We first tried to use GILL \cite{kohGroundingLanguageModels2023} directly with the default settings other than the maximum number of output tokens being set to $512$ so as to be an appropriate length for a full article. We enable both generation and retrieval for GILL. We experimented with prompts such as ``Please write me a step-by-step article for the goal {GOAL}" and many variants of it. Despite this, we found that the generations were always substantially shorter than a full article, likely due in part to the short length of the training data GILL has seen. We experimented with different values for the temperature as well as other parameters, but were unable to produce longer outputs that would adequately illustrate a goal. The text quality of these outputs was also very low. For instance, when prompted about the goal ``How to Make Apple Pie,'' GILL responded ``I can’t imagine why you would want to do this.''.  Typically, the outputs would comprise text for a single step or a single image, and we did not observe any instances of multiple.

\begin{figure}[h!]
    \centering
    \includegraphics[width=\linewidth]{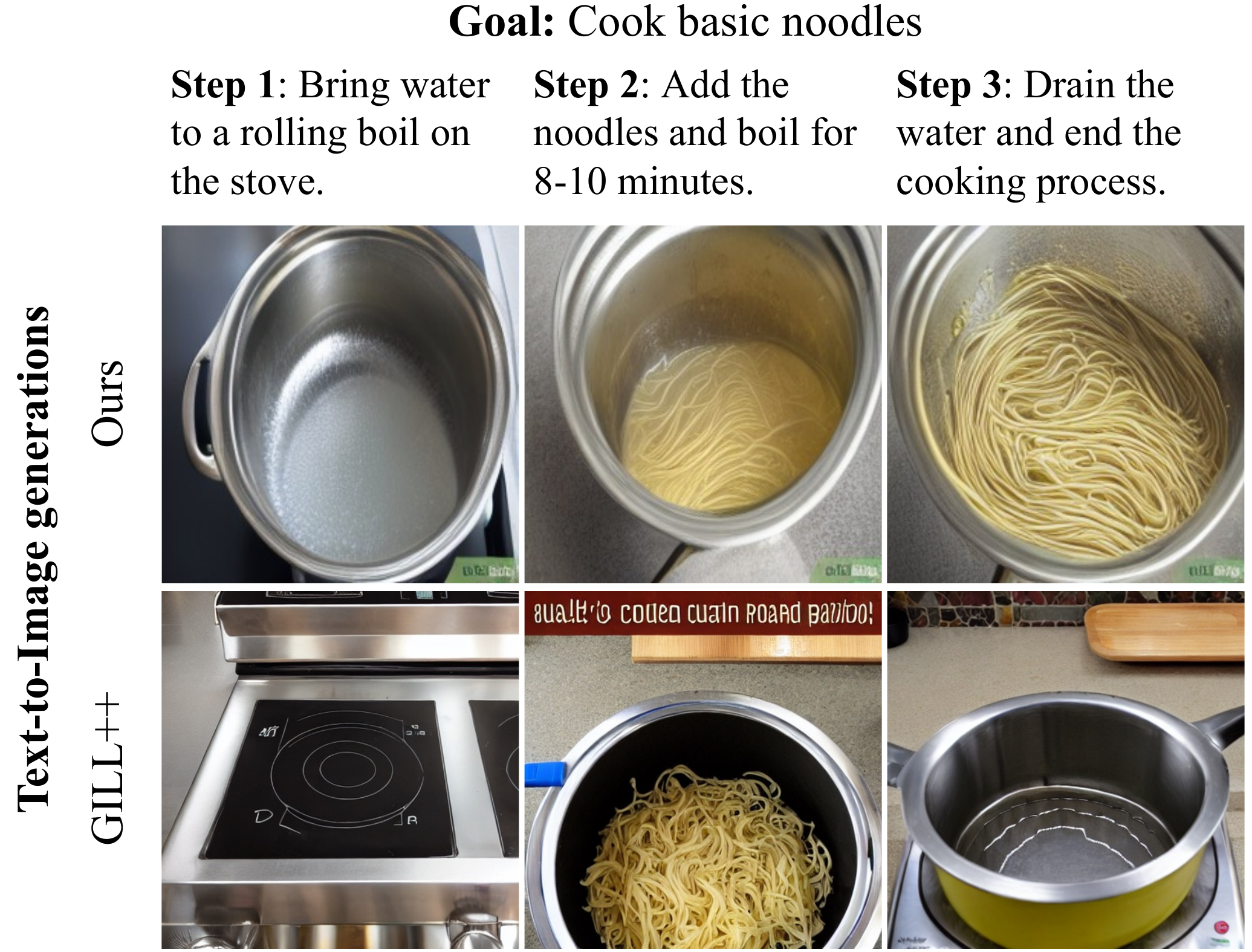}
    \caption{{\bf Comparison with our GILL++ .}}
    \label{fig:gill}
  \end{figure}

Because of this, we elected to try to surmount the language modeling difficulties of GILL and see how well it can perform when given text from the same LLM used for our experiments, which we call ``GILL++". This would evaluate the image generation and retrieval capabilities of GILL on their own. (To some extent, this does defeat the purpose of a multimodal generative model; it reflects the limitations of current generative models.) We performed LLM inference as described in Appendix \cref{sec:appdx:promptengg}. We provided the generated text to GILL with the prompt ``Please illustrate the step: \{STEP\} for the goal: \{GOAL\}''. We also experimented with various values of the ``sf'' (scaling factor) inference parameter, which the GILL codebase comments: ``increase to increase the chance of returning an image''. However, the model did not always produce an image even when increasing this value. The eventual parameters we arrived at for GILL++ were a scaling factor of $1.4$, temperature of $0.6$, $top\_p$ of $0.95$, and all other parameters left the same. A qualitative comparison can be seen in Figure \ref{fig:gill}. We find that the image generations still remain inferior to ours even when the text is fixed.

\section{Retrieval Baselines}\label{sec:appdx:retrieval}
We implement two separate retrieval-based baselines that use real images for the illustrations of each step. We describe these both below.

\subsection{LLM + CLIP}
The LLM+CLIP baseline is a combination of text generated by an LLM and real images retrieved from that text. We prompt the LLM the same way as for the final \METHODNAME model. After obtaining these steps, we retrieve the top image from the training dataset using CLIP similarity. For each step, we rank all images in the training set by their CLIP similarity to that step's text, and use the most similar image for each. Each of these images can come from any disparate goal in the training set.

\subsection{Goal Retrieval}
For the goal retrieval baseline, we pick the closest full article from the training set to compare to. The intent of this baseline is to show that the model does not simply reproduce the closest article it saw during training but rather generates new articles. We use the CLIP text similarity compare the generated goal text to the training set. We then pick the closest article to the generated goal text. We use the steps from this article as the steps for the generated article. This baseline is similar to the LLM + CLIP baseline, but instead of using the LLM to generate the steps, we use the steps from the closest article in the training set and simply use the images associated with those ground-truth steps.

\section{Additional Qualitative Results}
\begin{figure}[t]
    \centering
    \includegraphics[width=\linewidth]{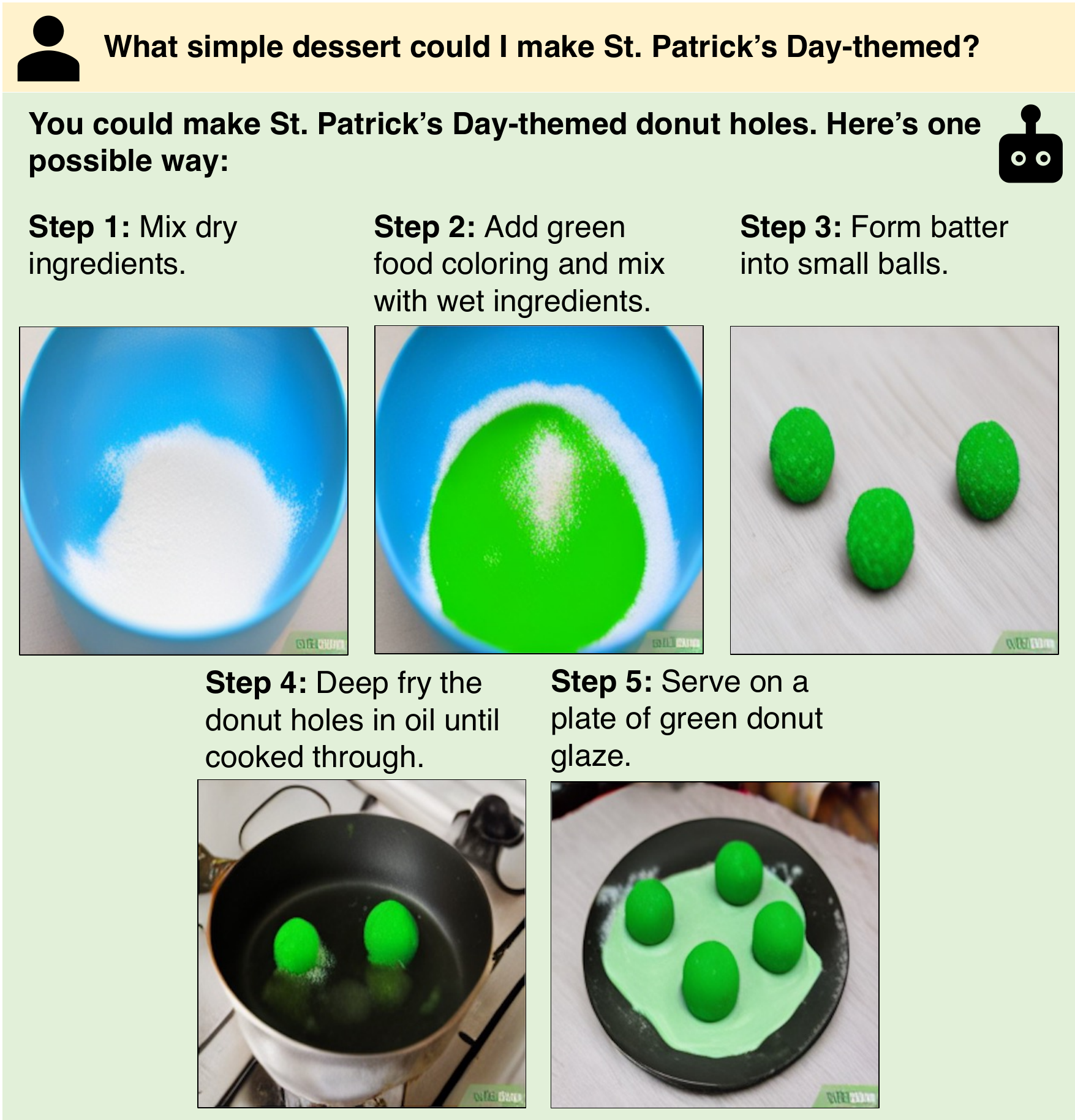}
    \caption{{\bf Additional personalization.} The steps illustrate donut holes customized to be green for St. Patrick's day.}
    \label{fig:stpatricks}
  \end{figure}

\begin{figure}[t]
      \centering
      \includegraphics[width=\linewidth]{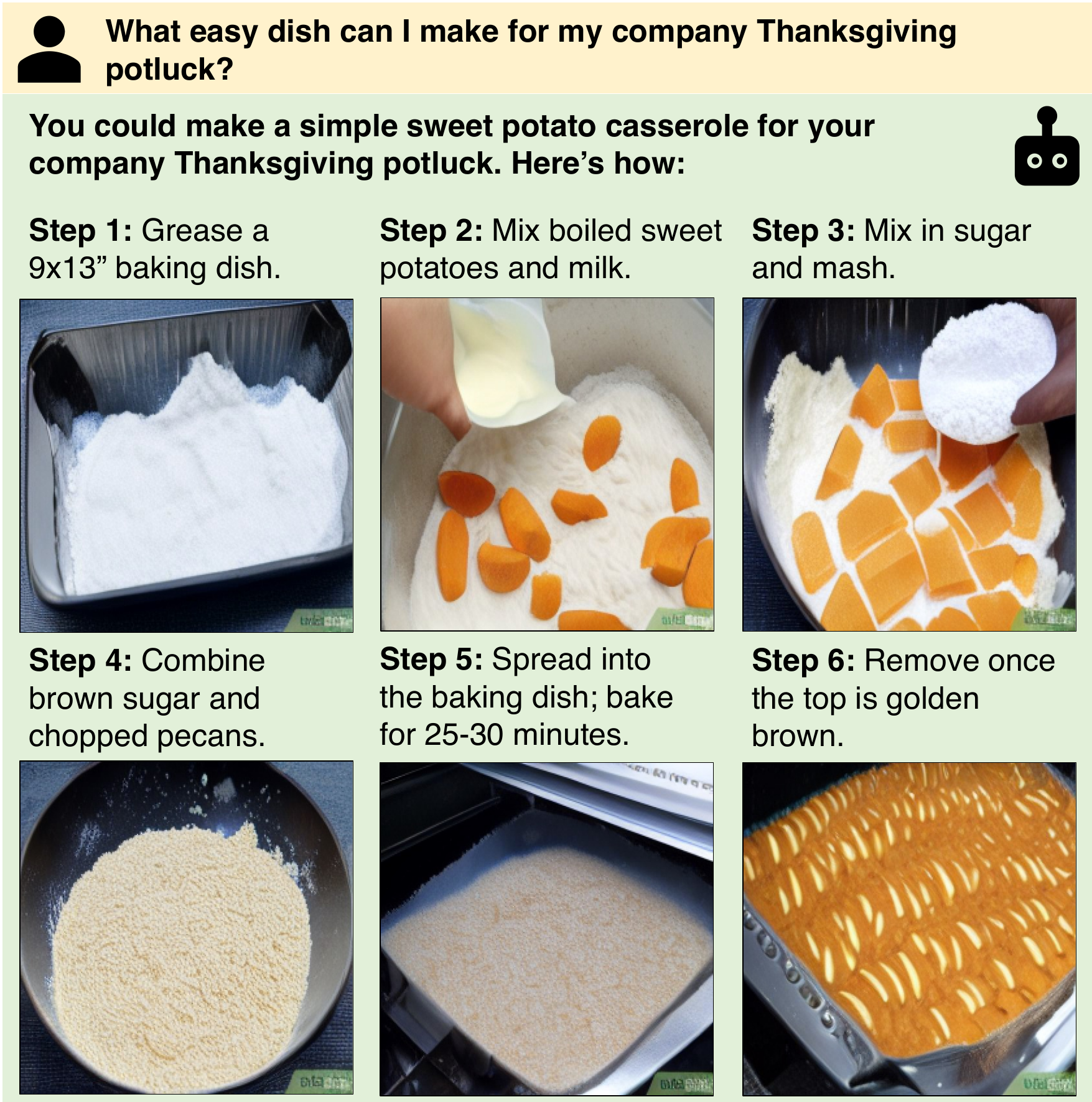}
      \caption{{\bf Additional goal suggestion.} \METHODNAME suggests the goal of making a casserole and provides illustrated instructions for it.}
      \label{fig:thanksgiving}
    \end{figure}

\begin{figure}[t]
      \centering
      \includegraphics[width=\linewidth]{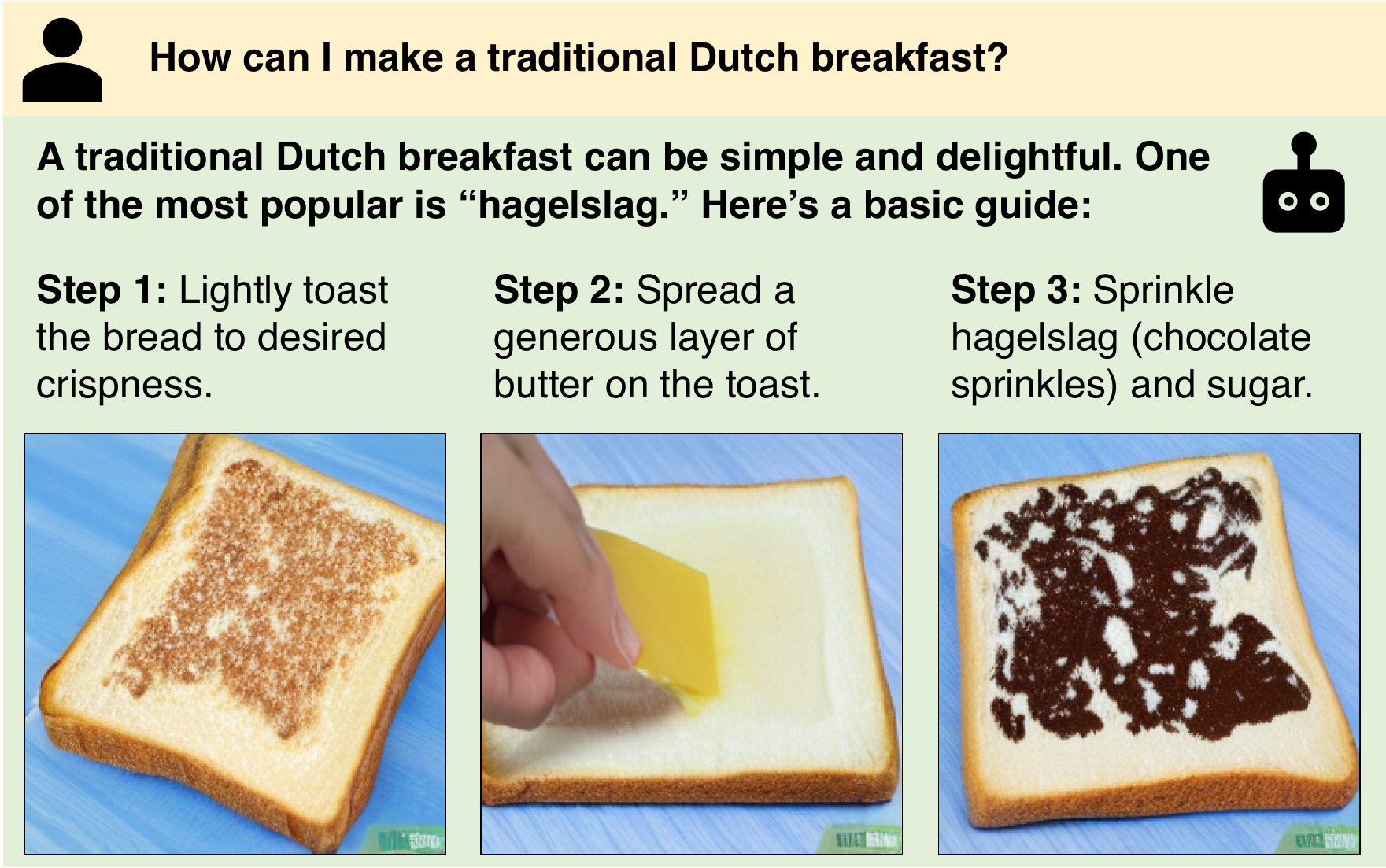}
      \caption{{\bf Additional goal suggestion.} The world knowledge of the LLM allows it to inform the user about hagelslag.}
      \label{fig:hagelslag}
    \end{figure}

\begin{figure}[t]
      \centering
      \includegraphics[width=\linewidth]{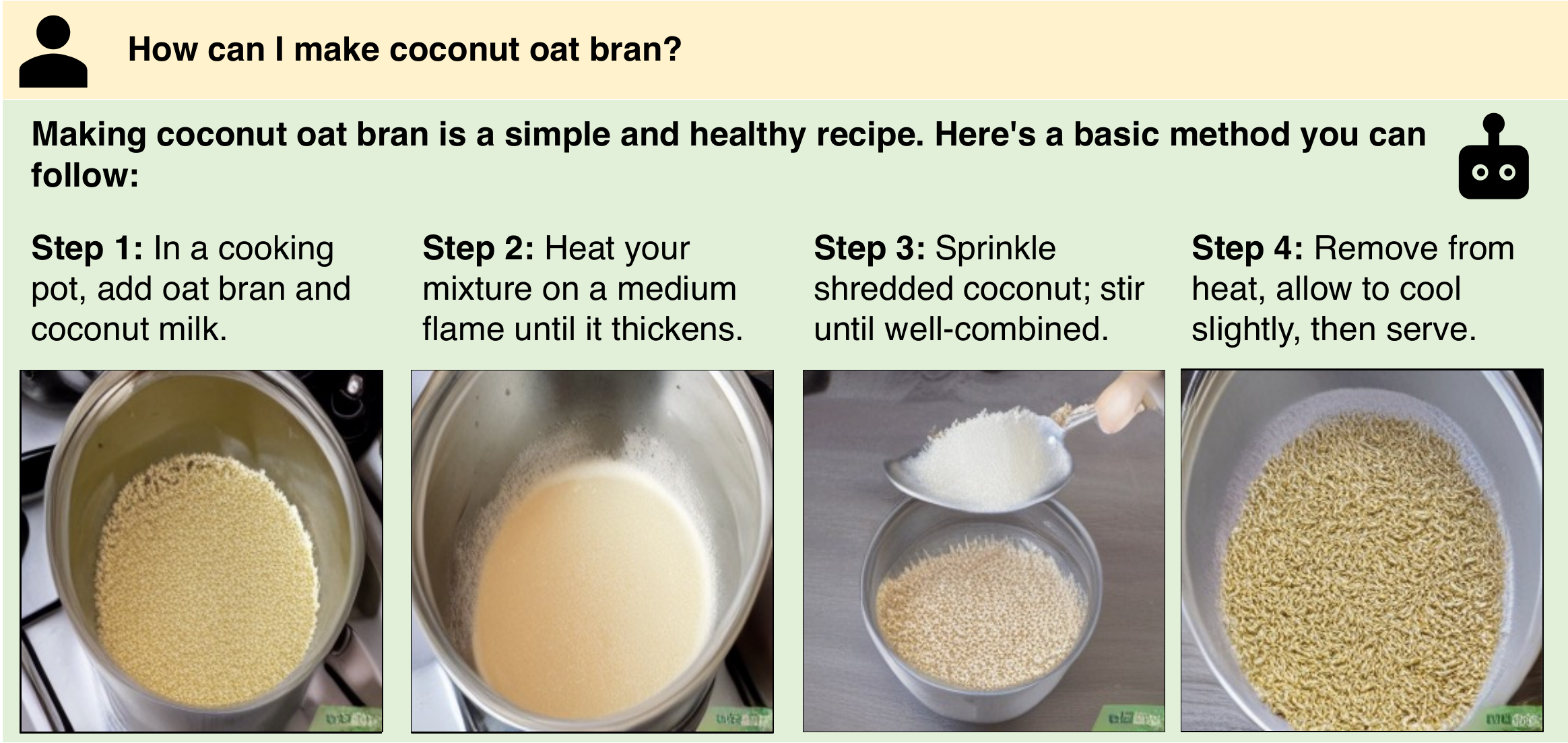}
      \caption{{\bf Additional article generation.} Coconut oat bran is a variant on an existing dish (not found in the training set) that \METHODNAME can illustrate.}
      \label{fig:oatbran}
    \end{figure}

\begin{figure}[t]
      \centering
      \includegraphics[width=\linewidth]{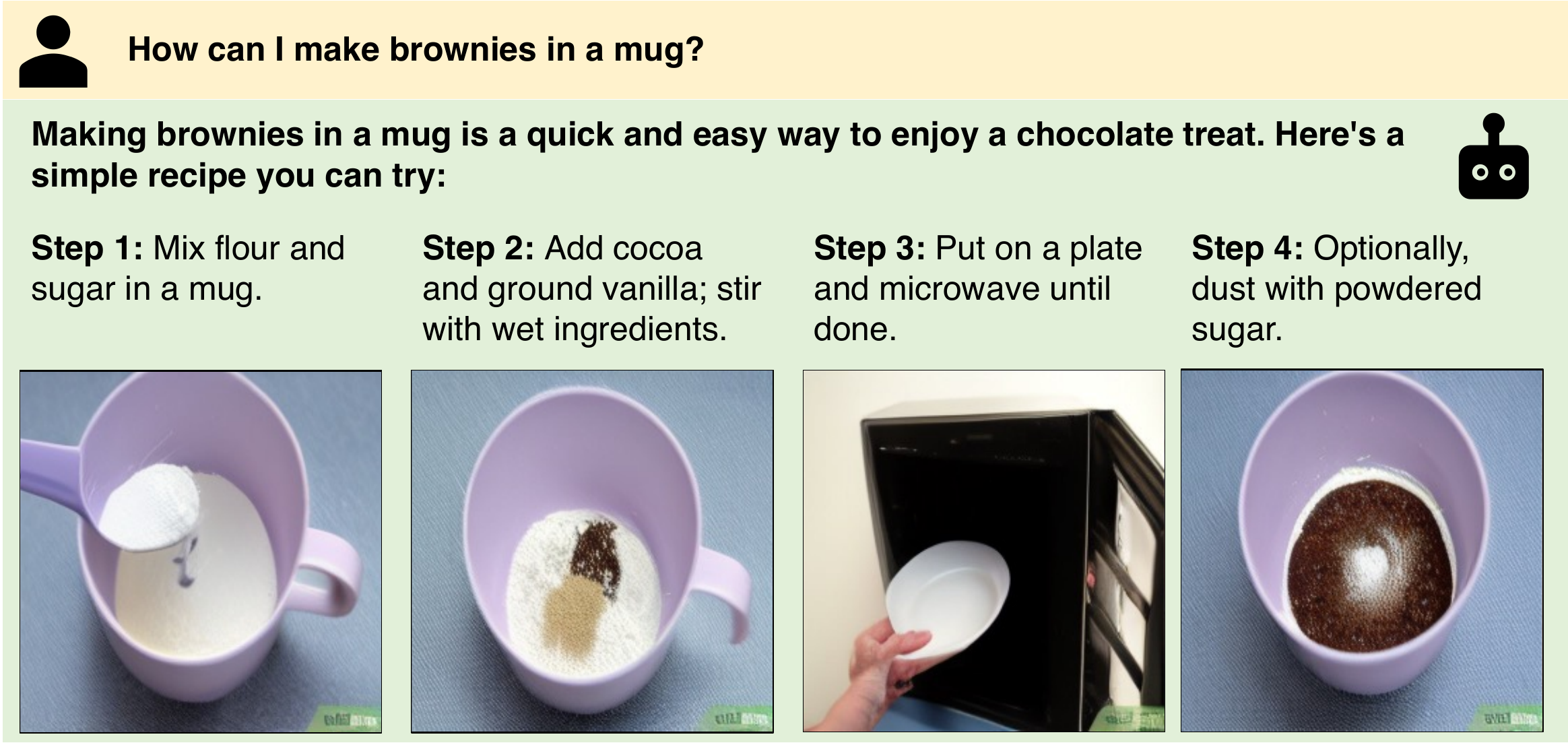}
      \caption{{\bf Additional article generation.} The mug shows the consistency \METHODNAME enables.}
      \label{fig:mugbrownie}
    \end{figure}

\begin{figure}[h]
      \centering
      \includegraphics[width=\linewidth]{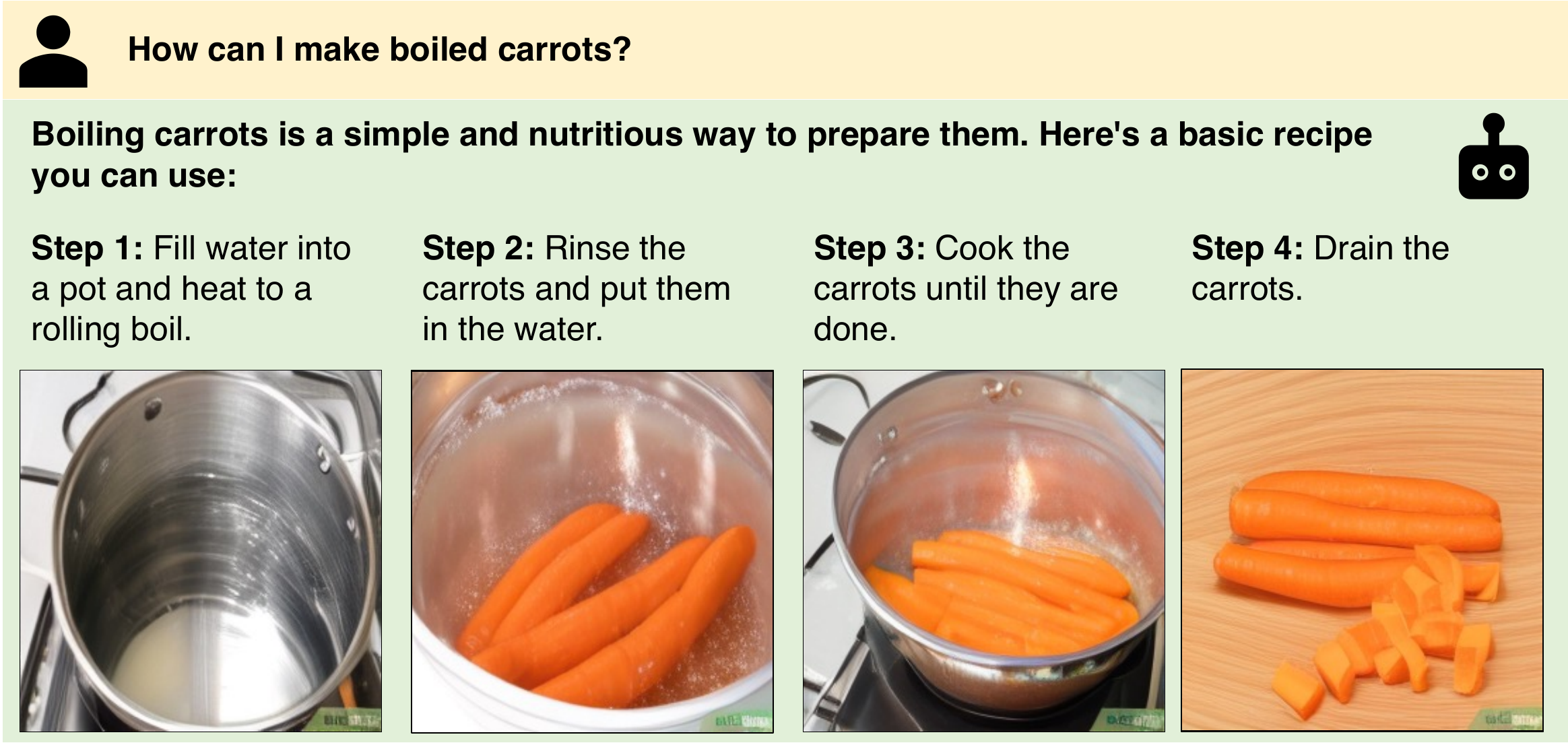}
      \caption{{\bf Additional article generation.}}
      \label{fig:carrots}
    \end{figure}

\begin{figure}[h]
      \centering
      \includegraphics[width=\linewidth]{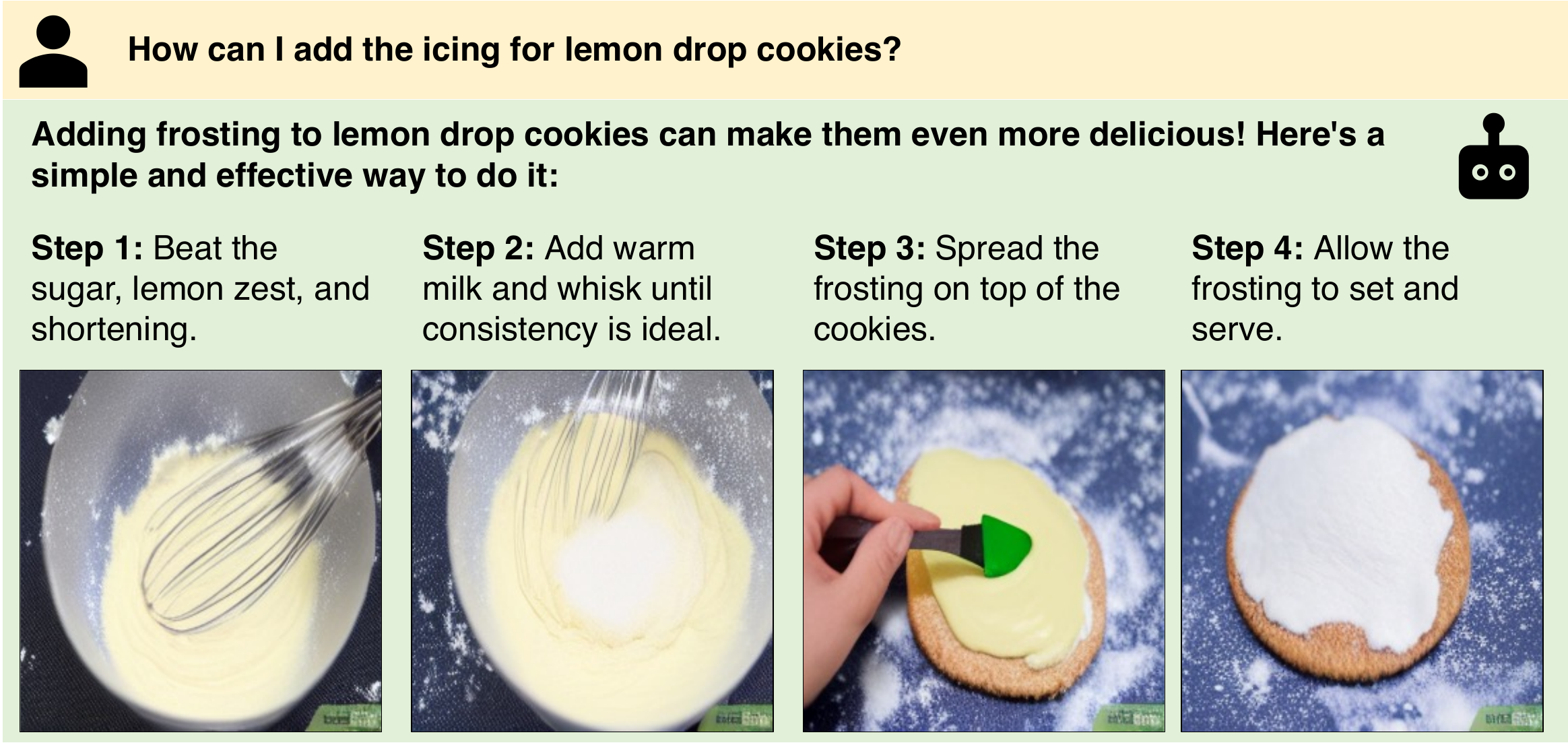}
      \caption{{\bf Additional article generation.} }
      \label{fig:lemondrop}
    \end{figure}

\begin{figure}[h]
      \centering
      \includegraphics[width=\linewidth]{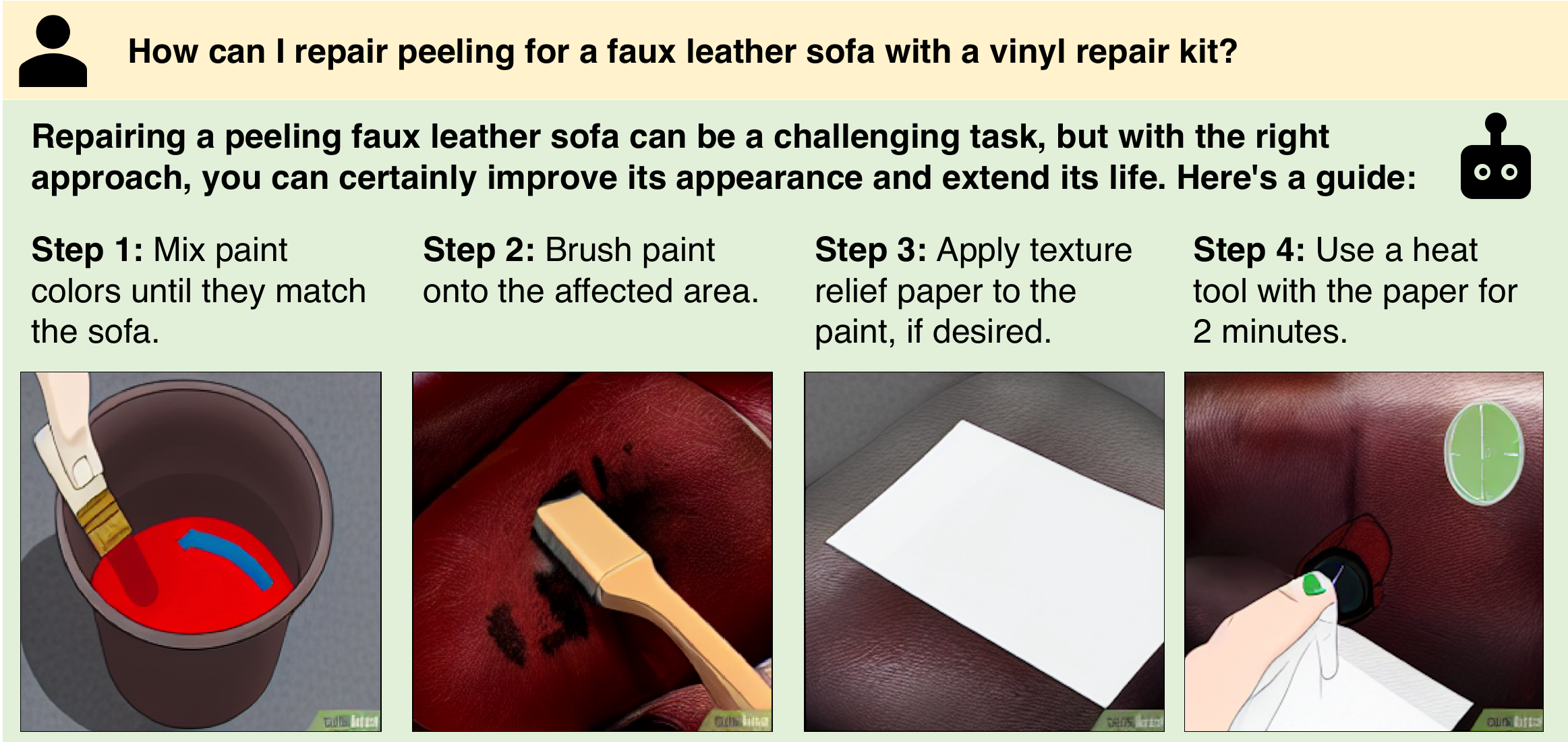}
      \caption{{\bf Additional article generation outside of recipes.} }
      \label{fig:repairsofa}
    \end{figure}

We present additional qualitative results in \cref{fig:stpatricks}, \cref{fig:thanksgiving}, \cref{fig:hagelslag}, \cref{fig:oatbran}, \cref{fig:mugbrownie}, \cref{fig:carrots}, \cref{fig:lemondrop}, and \cref{fig:repairsofa}. In particular, we note that \cref{fig:repairsofa} shows that \METHODNAME can generate articles outside of the recipe domain, such as for furniture repair.

\section{Failure Cases}\label{sec:appdx:failures}

\begin{figure}[t]
      \centering
      \includegraphics[width=\linewidth]{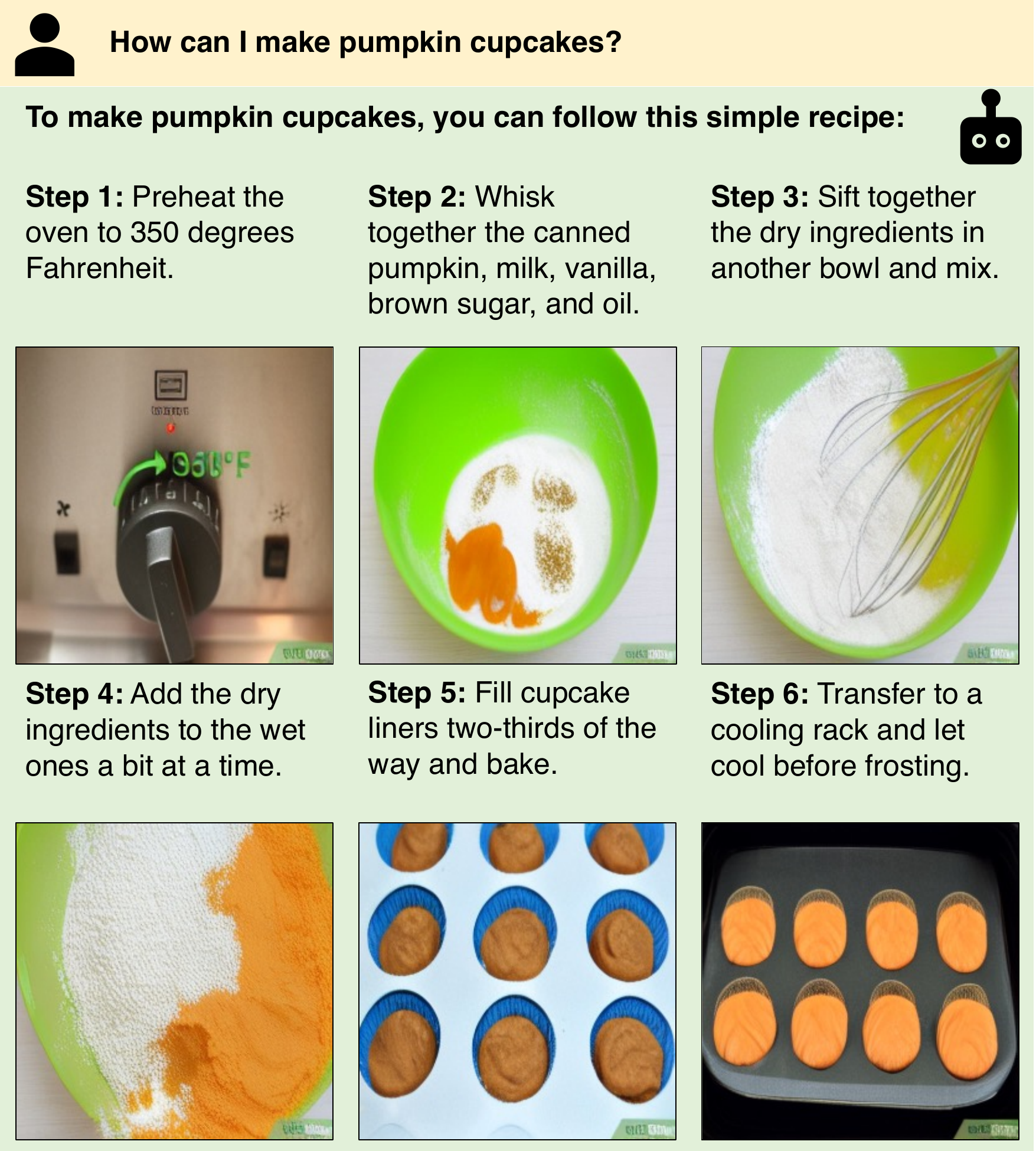}
      \caption{{\bf Example failure case.} We note two failure modes. First, the model is over-consistent, producing a bowl in Step 3 that looks similar to the bowl in Step 2 when it should be another bowl. Second, the model appears to consider geometric arrangements as similar to each other even if the numbers in each differ, leading to inconsistency between Step 5 and Step 6.}
      \label{fig:pumpkin_fail}
    \end{figure}

\begin{figure}[t]
      \centering
      \includegraphics[width=\linewidth]{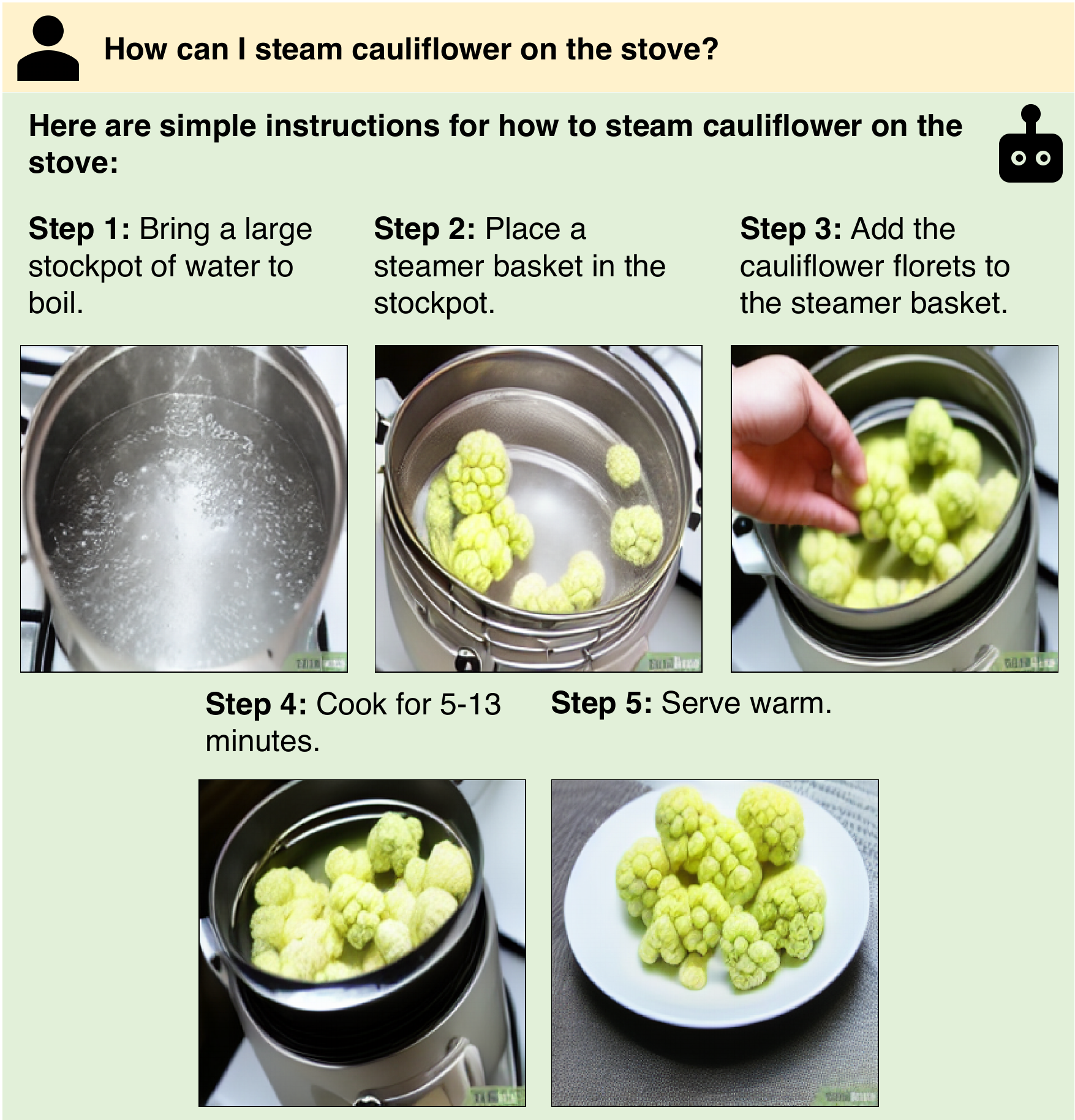}
      \caption{{\bf Example failure case 2.} An additional example displaying over-consistency at the cost of step faithfulness. Step 2 already displays cauliflower in the in the steamer basket, but the text only describes it being added in Step 3.}
      \label{fig:cauli_fail}
    \end{figure}

In addition, we present examples of failure cases in \cref{fig:pumpkin_fail,fig:cauli_fail}. We observe one failure mode in being \textit{over-consistent}: while Step 3 says to mix in `another' bowl, the model ends up producing a bowl that looks similar to the bowl in Step 2. We also observe a case where the model finds consistency difficult to produce: geometric arrangements seem to be considered similarly, making a 3x3 grid of cupcakes into a 2x4 grid.


\end{document}